\documentclass[11pt]{article}

\usepackage[preprint]{acl}
\usepackage{hyperref}
\usepackage{comment}   % in preamble
\usepackage{xurl}
\usepackage{comment}
\usepackage[table]{xcolor}
\usepackage{array}
\usepackage{times}
\usepackage{latexsym}
\usepackage{multirow}
\usepackage{xcolor}
\usepackage{longtable}
\usepackage{dblfloatfix}
\usepackage{caption}
\usepackage[T1]{fontenc}
\usepackage[utf8]{inputenc}

\usepackage{microtype}

\usepackage{inconsolata}

\usepackage{graphicx}
\usepackage{makecell}
\usepackage{amsmath}
\usepackage{amssymb}
\usepackage{mathtools}
\usepackage{amsthm}
\usepackage{bbding}

\usepackage{makecell}  % for better cell formatting
\usepackage{adjustbox}

\usepackage{subcaption}
\usepackage{tabularx}
\usepackage{dsfont}
\usepackage{multirow}
\usepackage{wrapfig}
\usepackage{siunitx}
\usepackage{amssymb}
\usepackage{multirow}
\usepackage{color}
\usepackage[capitalize,noabbrev]{cleveref}
\usepackage{lipsum,booktabs}
\usepackage{natbib}

\usepackage{multicol}

\usepackage{caption}
\usepackage{subcaption}
\usepackage{enumitem}

\usepackage{upgreek}

\usepackage{float}

\usepackage{array}

\title{{\textbf{\textsc{LoWAgent}}: A Low-Resource Multilingual Benchmark for Web Agents }}

\author{
\textbf{Ocean Monjur$^{1}$, Daud Ahmad Nisar$^{1}$, Shahriar Kabir Nahin$^{1}$, Tadesse Destaw Belay$^{2}$} \\
\textbf{Sushruti Mishra$^{3}$, Israel Abebe Azime$^{4}$, Anshuman Chhabra$^{1}$} \\
    $^{1}$University of South Florida, USA \quad $^{2}$Instituto Politécnico Nacional, Mexico \\
    $^{3}$Independent Researcher \quad $^{4}$Saarland University, Germany
}

\begin{document}
\maketitle
\makeatletter
{\renewcommand\thefootnote{}\long\def\@makefntext#1{\noindent #1}\footnotetext{Correspondence: \texttt{\{omonjur,anshumanc\}@usf.edu}}}
\makeatother

%\author{
%  \textbf{First Author\textsuperscript{1}},
%  \textbf{Second Author\textsuperscript{1,2}},
%  \textbf{Third T. Author\textsuperscript{1}},
%  \textbf{Fourth Author\textsuperscript{1}},
%\\
%  \textbf{Fifth Author\textsuperscript{1,2}},
%  \textbf{Sixth Author\textsuperscript{1}},
%  \textbf{Seventh Author\textsuperscript{1}},
%  \textbf{Eighth Author \textsuperscript{1,2,3,4}},
%\\
%  \textbf{Ninth Author\textsuperscript{1}},
%  \textbf{Tenth Author\textsuperscript{1}},
%  \textbf{Eleventh E. Author\textsuperscript{1,2,3,4,5}},
%  \textbf{Twelfth Author\textsuperscript{1}},
%\\
%  \textbf{Thirteenth Author\textsuperscript{3}},
%  \textbf{Fourteenth F. Author\textsuperscript{2,4}},
%  \textbf{Fifteenth Author\textsuperscript{1}},
%  \textbf{Sixteenth Author\textsuperscript{1}},
%\\
%  \textbf{Seventeenth S. Author\textsuperscript{4,5}},
%  \textbf{Eighteenth Author\textsuperscript{3,4}},
%  \textbf{Nineteenth N. Author\textsuperscript{2,5}},
%  \textbf{Twentieth Author\textsuperscript{1}}
%\\
%\\
%  \textsuperscript{1}Affiliation 1,
%  \textsuperscript{2}Affiliation 2,
%  \textsuperscript{3}Affiliation 3,
%  \textsuperscript{4}Affiliation 4,
%  \textsuperscript{5}Affiliation 5
%\\
%  \small{
%    \textbf{Correspondence:} \href{mailto:email@domain}{email@domain}
%  }
%}

\begin{abstract}
\looseness-1 Web agents are typically evaluated on English-language interfaces, leaving their ability to operate across the world’s multilingual websites largely unexplored. This gap is especially significant for \textit{low-resource languages}, where standardized resources for evaluating web agents remain limited. To address this gap, we introduce \textit{\textbf{Lo}w-Resource \textbf{W}eb \textbf{Agent} Benchmark} (\textbf{\textsc{LoWAgent}}), a benchmark for evaluating web agents across different low-resource languages (LRLs) in comparison to the default, English. \textsc{LoWAgent} evaluates web agents across English and four LRLs: \textit{Hindi, Bengali, Urdu}, and \textit{Amharic}. The benchmark comprises of six fully localized websites, instantiated across five languages to yield 3,000 task instances. We evaluate three closed-source and five open-source models across all five languages on all six websites. Our findings show that both closed-source and open-source models experience performance degradation for LRLs compared to English, with substantially greater performance drops observed among the open-source model agents. We further evaluate the robustness of the models on \textit{realistic interface disruptions}, such as \textit{cookie notices} and \textit{advertisement pop-ups}, and find that these further degrade agentic performance. Finally, we show that post-training on a small subset of \textsc{LoWAgent} traces can significantly improve low-resource language agentic task performance, demonstrating the usefulness of our benchmark.

\end{abstract}

\section{Introduction}
\label{sec:intro}

\looseness-1 Large Language Models (LLMs) \cite{brown2020language} have driven significant advances in AI agents, enabling them to interact with external environments and autonomously execute increasingly complex, multi-step tasks such as software engineering, web navigation, scientific discovery, data analysis, and embodied robotic control \cite{zhang2025reinforcement,haase2026beyond,ghafarollahi2025sciagents}. In particular, \textit{web agents} have emerged as a promising paradigm for leveraging LLMs to navigate websites.
As these agents become increasingly capable, reliably evaluating their ability to operate in realistic web environments has emerged as an important research problem \cite{zhou2024webarena,he2024webvoyager}. 

Prior work has introduced benchmarks for evaluating web agents across diverse websites and task settings, including web navigation, information retrieval, planning, and interaction with dynamic interfaces~\cite{zhou2024webarena,koh2024visualwebarena,he2024webvoyager}. Other work has begun to extend web agent evaluation to the multilingual setting. X-WebAgentBench \cite{wang2025x} evaluates web agents across 14 languages only within a single shopping environment, while WebMMU \cite{awal2025webmmu} explores multimodal and multilingual website understanding across four high-resource languages (English, French, German, and Spanish). However, existing web agent benchmarks have predominantly focused on English and other high-resource languages, leaving the performance of low-resource languages (LRLs), largely unexplored. This limitation is especially important because prior work has shown that LLM performance can vary significantly on tasks involving LRLs~\citep{dey2024better,hofman2026maps}.

\begin{figure*}[htb!]
    \centering
    \includegraphics[width=.99\textwidth]{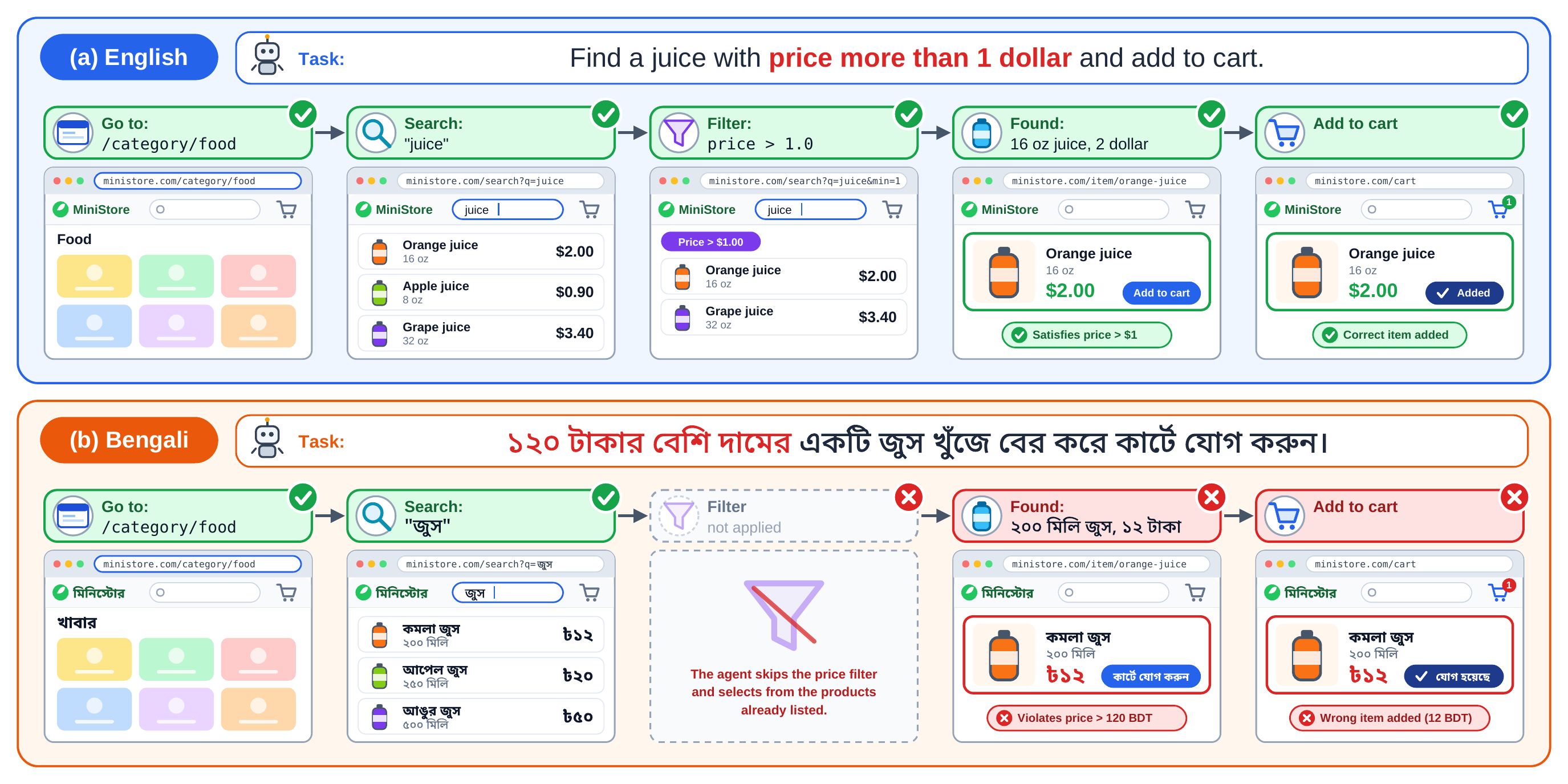} %\vspace{-3mm}
    \caption{Example trajectories for equivalent shopping tasks in (a) \textit{English} and (b) \textit{Bengali}. The agent successfully completes the English task by applying the required price filter and selecting a valid item. In contrast, for the Bengali task, the agent correctly navigates and searches for the product but skips the filtering step, resulting in the selection of an item that violates the specified price constraint and leading to task failure.} \vspace{-4mm}
    \label{fig:intro}
\end{figure*}

To address this gap in the evaluation of web agents for low-resource languages, we introduce \textit{\textbf{Lo}w-Resource \textbf{W}eb \textbf{Agent} Benchmark} (\textbf{\textsc{LoWAgent}}), a benchmark for evaluating web agents across five languages: \textit{English, Bengali, Hindi, Urdu}, and \textit{Amharic}. Here, Bengali, Hindi, Urdu, and Amharic are representative LRLs, while English serves as a reference point for comparison and provides a standard default baseline for evaluating web-agent performance. Additionally, the benchmark covers a diverse set of web environments, requiring agents to interact with different website layouts, functionalities, and task types. These tasks include (1) \textit{shopping}, (2) \textit{information seeking}, (3) \textit{booking}, and (4) \textit{social interactions}. We ensure the same website characteristics across all five languages. Our diverse evaluation settings enable a more comprehensive assessment of web agent capabilities and help us understand how language can influence web agent performance (see example in Figure \ref{fig:intro}).

\looseness-1We further use \textsc{LoWAgent} to evaluate both closed-source and open-source LLM agents, such as GPT-6 Luna, GPT-5.6 Luna, Gemini 3.1 Flash-Lite, Llama-3.1-8B, Gemma3 (4B and 12B variants), GPT-OSS-20B, and compare how their performance varies across English and LRL settings.  To make the evaluation more realistic, we also introduce interface-level disruptions that reflect common obstacles encountered in real-world web environments, such as cookie notices and advertisement pop-ups. Finally, we investigate whether post-training on a subset of \textsc{LoWAgent} data can improve performance on held-out tasks and whether these improvements hold across languages. Overall, \textsc{LoWAgent} covers a diverse range of LRLs, web environments, task types, and interface conditions, providing a comprehensive benchmark for evaluating web agents in realistic low-resource settings.\vspace{4mm}

\noindent\looseness-1 In sum, we make the following \textbf{contributions}:

\begin{itemize}[leftmargin=*, nosep]

\item We introduce \textsc{LoWAgent}, a multilingual agentic benchmark spanning five languages (4 LRLs and English) and six diverse websites, with 450 text-based and 150 multimodal tasks, similarly instantiated in all five languages.

\item For realism, we evaluate web agent robustness to common interface disruptions, such as cookie notices and advertisement pop-ups as well.

\item Furthermore, to mitigate performance issues, our experiments show that post-training on a subset of \textsc{LoWAgent} data substantially improves open-weight web agent performance across LRLs.

\end{itemize}

\section{Related Works}
With the rise of agentic systems and their growing ability to interact with the web \cite{shi2017world,liu2018reinforcement,deng2023mind2web}, recent work has increasingly focused on developing benchmarks to evaluate web agent capabilities. WebArena \cite{zhou2024webarena} was among the first benchmarks to evaluate web agents in realistic, interactive environments, comprising a diverse set of functional websites and long-horizon tasks. WebArena includes 812 tasks across five website environments: e-commerce, social forums, collaborative software development, content management, and maps. Building on WebArena, VisualWebArena \cite{koh2024visualwebarena} extends web agent evaluation to visually grounded interaction, introducing 910 tasks that require agents to reason over both textual and visual information. WebVoyager \cite{he2024webvoyager} instead evaluates web agents through end-to-end interaction with 15 real-world websites, using both screenshots and textual representations of web elements to guide agent actions. Mind2Web \cite{deng2023mind2web} broadens web agent evaluation to real-world websites, comprising over 2,000 tasks collected from 137 websites across 31 domains, while AssistantBench \cite{yoran2024assistantbench} evaluates agents on 214 realistic, time-consuming web tasks spanning diverse domains.

Prior work has explored LRL evaluations across a range of NLP tasks \cite{belay2026afrilangtutor,azime2026bridging}, with recent work beginning to extend web agent evaluation beyond English. X-WebAgentBench \cite{wang2025x} is a multilingual interactive web benchmark spanning 14 languages and 2,800 instructions within a single shopping environment. WebMMU \cite{awal2025webmmu} further explores multimodal and multilingual website understanding across four languages and more than 20 website domains, covering English, German, Spanish, and French, all of which are relatively high-resource languages. Despite these advances, systematic evaluation of web agents in low-resource languages across different website environments remains limited, which we investigate in this work.

\section{ \textsc{LoWAgent}}

In our work, we seek to investigate the following research questions on the performance of web agents in low-resource languages (LRLs) versus English:
\begin{itemize}[leftmargin=*, nosep]
\item \textbf{RQ1 (Multilingual Performance):} How does web agent performance in low-resource languages compare with English?

\item \textbf{RQ2 (Interface Disruptions):} To what extent do cookie notices and advertisement pop-ups affect web agent performance across LRLs?

\item \textbf{RQ3 (Performance Improvement):} Can post-training on a subset of \textsc{LoWAgent} data improve LLM performance across agentic tasks in LRLs?
\end{itemize}

To formally evaluate our RQs, we prepare \textsc{LoWAgent}, a benchmark for evaluating how web agent performance varies across languages. For comprehensive evaluation, we develop six functional websites, each localized into five languages: English (EN), Bengali (BN), Hindi (HI), Urdu (UR), and Amharic (AM). For each language, we create 75 distinct text-based tasks for each website, yielding 450 unique tasks across the six websites and 2,250 task instances across all five languages. To evaluate agents on visually grounded tasks, we additionally include 50 such tasks for three of the six websites, resulting in 150 unique multimodal tasks and 750 task instances across the five languages. Combined with the text-based tasks, \textsc{LoWAgent} contains 600 unique tasks and 3,000 total task instances. Appendix \ref{Appendix: website} provides further details about each website and its functionalities.

We design the six websites to cover diverse web agent scenarios, including an e-commerce store (Ministore), discussion forum (Redora), used-car marketplace (GariBazaar), railway ticketing system (RailSetu), government services portal (Haritia National Portal), and airline booking platform (Selam Airways). Three websites, namely Ministore, Redora, and GariBazaar, additionally include visually grounded tasks that require agents to reason over visual content. For each website, we maintain the same codebase, database, task identifiers, and ground-truth states across the five language versions, varying only the language used.

\subsection{Task Formulation}
Following \citet{zhou2024webarena}, we model each website in language $\ell$ as an environment $\mathcal{E}_\ell=\langle \mathcal{S_\ell},\mathcal{A_\ell},\mathcal{O_\ell},\mathcal{T_\ell}\rangle$ with state space $\mathcal{S_\ell}$, action space $\mathcal{A_\ell}$, observation space $\mathcal{O_\ell}$, and a deterministic transition function $\mathcal{T_\ell}:\mathcal{S_\ell}\times\mathcal{A_\ell}\rightarrow\mathcal{S_\ell}$. The state comprises the site's database together with the agent's current page and session state. Given a natural-language intent $i$ written in $\ell$, at step $t$ the agent receives an observation $o_{\ell,t}\in\mathcal{O_\ell}$ and issues an action $a_{\ell,t}\in\mathcal{A_\ell}$ conditioned on $i$, $o_{\ell,t}$, and its last $k$ steps. The environment transitions to the next state according to $s_{\ell,t+1}=\mathcal{T_\ell}(s_{\ell,t},a_{\ell,t})$, where the resulting state reflects the effect of action $a_{\ell,t}$ on the current environment state $s_{\ell,t}$. An episode terminates when the agent stops or reaches the step budget. A binary reward function $r(i,a_{\ell,T_\ell},,s_{\ell,T_\ell})\in\{0,1\}$ then scores the episode based on the agent's final answer, if applicable, and the final state $s_{\ell,T_\ell}$.

% \subsection{Websites}
% \textsc{LoWAgent} contains six websites from domains common in everyday web use in South Asia and East Africa: an online store, a discussion forum, a used-car marketplace, a railway-ticketing site, a government-services portal and an airline (Table \ref{tab:sites}).

\subsection{Observation Space}
\looseness-1 Agents interact with the web through a text-based browser that retrieves pages over HTTP and parses their HTML content without rendering pages visually. Each observation consists of (1) the page URL and title, (2) a numbered list of links, and (3) a numbered list of forms, including their fields, current values, available options, and submission buttons. For visual tasks, each observation also includes the task's reference image and a full-page screenshot, with links and forms numbered to match the text observation.

\begin{table}[t]
\centering
\small
\renewcommand{\arraystretch}{1.15}
\caption{Action space: List of all actions available to the web agent in \textsc{LoWAgent}.}\vspace{-3mm}
\begin{tabular}{lp{0.48\columnwidth}}
\toprule
\textbf{Action} & \textbf{Description} \\
\midrule
\texttt{goto <path>} & Opens a path on the current site. \\
\texttt{click <n>} & Follows link $n$. \\
\texttt{fill <n> k=v;} & Sets and submits the named fields of form $n$. When no fields are present, presses the form's button. \\
\texttt{stop <answer>} & Ends the episode with an answer. \\
\bottomrule
\end{tabular}
\vspace{-5mm}
\label{tab:actions}
\end{table}

\subsection{Action Space}
We define a compact action space that captures the core operations required for interacting with the websites. As summarized in Table~\ref{tab:actions}, the agent can navigate to a URL, follow links, fill and submit forms, or terminate an episode. Each action is executed atomically, with exactly one action issued per step. Form completion and submission are combined into a single action, allowing the agent to complete multi-field forms, such as checkout and passenger-information forms, in one step. Each action operates on the numbered elements in the agent's observation. For example, \texttt{click} selects the link at index $n$, while \texttt{fill k=v;} fills the specified fields of form $n$ and submits it. Any unspecified fields retain their existing values. 

For visual tasks, the action space is unchanged. Instead, each observation additionally includes a screenshot of the full page, in which every link and form is marked with the same index as in the text observation, so the agent can refer to what it sees through the existing actions. When a task requires an answer, the agent can end the episode with \texttt{stop}, optionally providing the answer. The browser environment executes and validates each action before returning the next observation. Actions that cannot be executed, such as clicking a nonexistent element, and outputs that fall outside the defined action space both result in an error message that is shown to the agent at the next observation.

\subsection{Multilingual Localization}
\label{sec:l10n}

As we have mentioned in previous sections, each website in \textsc{LoWAgent} is constructed in English, Bengali, Hindi, Urdu, and Amharic. To ensure comparability across languages, we preserve the underlying website functionality while localizing the user-facing content and task instructions. We describe the multilingual localization process and design choices of \textsc{LoWAgent} below.

\begin{figure*}[!t]
    \centering
    \includegraphics[width=.96\textwidth]{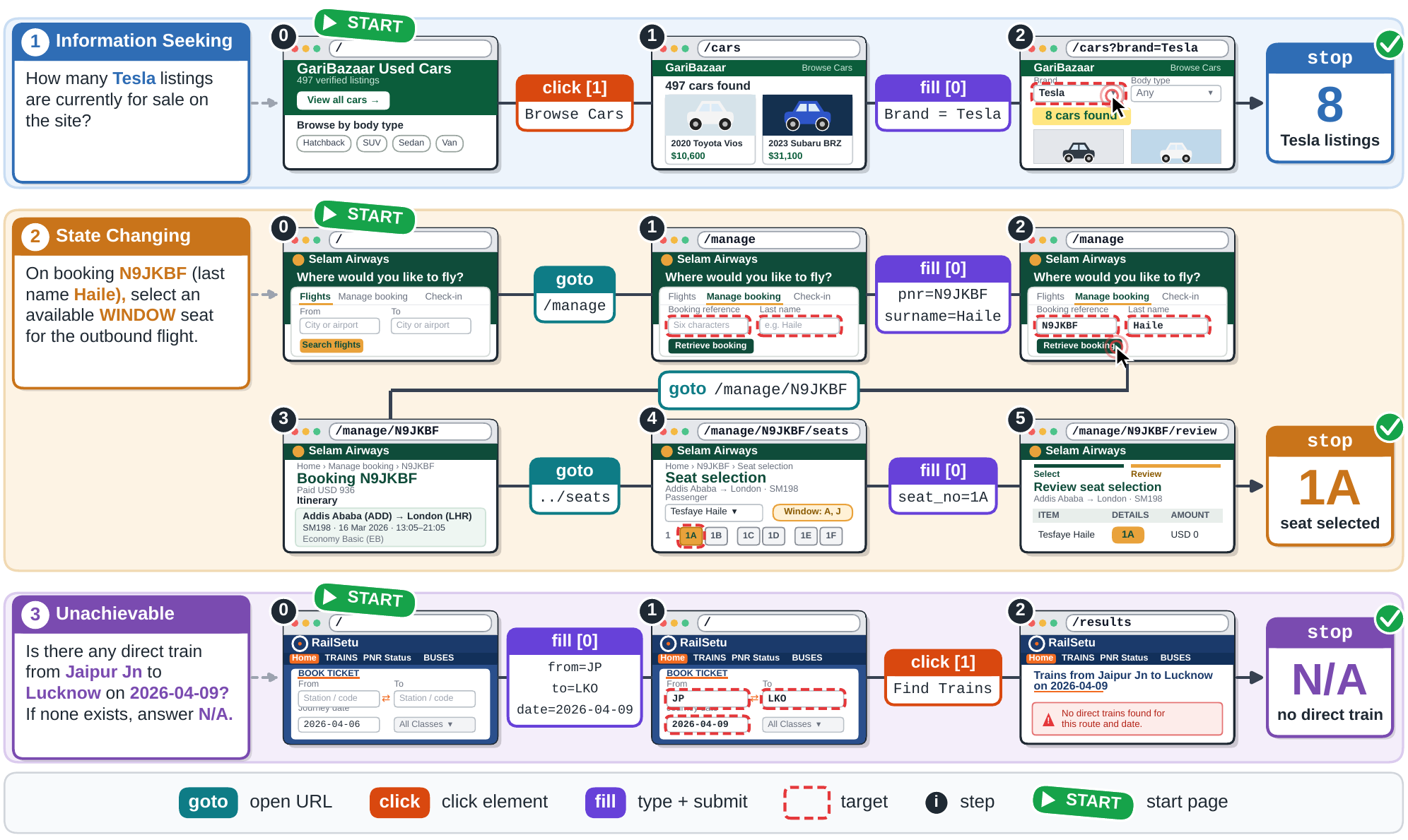}\vspace{-3mm}
    \caption{Example execution trajectory of the three different types of tasks in \textsc{LoWAgent}: (1) Information-seeking, (2) State-changing, and (3) Unachievable tasks.}\vspace{-5mm}
    \label{fig:Task example}
\end{figure*}

\looseness-1 For each website, the five language versions run on the same codebase. We localize the interface text, website content (e.g., person names, product names, posts, comments, listings, and service pages), and task intents while preserving identifiers and values that must remain consistent across languages. These include task IDs, product and listing IDs, PNRs, booking references, usernames, passwords, coupon codes, brand names, ISO-formatted dates, status codes, and the \texttt{N/A} answer token. Under this design, a task targets the same item and ground-truth state in every language, enabling direct comparison of agent performance across languages.

\vspace{0.5mm}

\noindent\textbf{Translation and Localization Process.}
We perform localization at the data and interface levels rather than maintaining separate implementations for each website. Interface text, including button labels, menus, and messages, is stored separately for each language. We use Fable 5.1 \cite{Fable} to assist with translating website content and interface text, followed by manual review and task-specific validation by \textbf{native speakers} to ensure consistency between localized content and the corresponding website state. Names of people and places are transliterated where appropriate rather than translated. Additionally, we generate task intents from language-specific templates using values from the corresponding localized website, ensuring that each intent refers to entities exactly as they appear in the agent's environment.

\vspace{0.5mm}

\looseness-1 \noindent\textbf{Scripts, Numerals, and Currency.}
We localize each website according to the writing conventions of its target language. Bengali and Hindi \cite{maung2025hybrid,arora2024devanagari} use their respective scripts, Urdu is presented in a right-to-left Nastaliq script \cite{naz2016urdu}, and Amharic uses the Ge'ez script \cite{gebremedhin2020linguistic}. While numerical values are stored in a standardized ASCII format, their display follows the numeral conventions of each localized interface. Where supported, input fields accept both localized numerals and ASCII digits. Prices on the websites are converted to the corresponding local currency using a fixed language-specific exchange rate, and price-dependent task answers are recomputed to preserve semantic equivalence across languages.

\subsection{Task Creation}
Each website contains 75 text-based tasks. We first manually design 30 task intents per website to cover diverse interaction patterns and task types. We use Fable 5.1 to expand a set of human-written tasks (30) into a larger pool of task templates (75 for text and 50 for visual tasks), where each template is a parameterized task with placeholder slots. A programmatic generator then instantiates each template into concrete tasks by filling those slots with values drawn from the website's underlying data. Each task consists of a natural-language intent, a starting page, a step budget of 30 or 40 actions (based on the task), and a reference action sequence. Finally, each task is manually executed and reviewed by human verifiers to verify that it is valid, unambiguous, and successfully solvable in the corresponding website environment.
\vspace{0.5mm}

\looseness-1 \noindent\textbf{Task Types.}
Following \citet{zhou2024webarena}, we categorize tasks into three types: (1) \emph{information-seeking}, (2) \emph{state-changing}, and (3) \emph{unachievable}. As demonstrated in Figure \ref{fig:Task example}, information-seeking tasks require agents to retrieve or compute an answer from the website, while state-changing tasks require agents to perform an action that results in a target database state. Unachievable tasks require the agent to recognize that the requested information or action is not possible, with \texttt{N/A} serving as the correct response. Additionally, we design these tasks to cover a range of complexity, from simple information lookups to multi-step workflows involving information retrieval, item selection, and transactional actions.

% Figure \ref{fig:Task example} shows examples of each task type.
\vspace{0.5mm}

\noindent\textbf{Visually Grounded Tasks.}
The e-commerce store, discussion forum, and used-car marketplace contain visually grounded tasks (50 per website and language). Following \citet{koh2024visualwebarena}, these tasks provide a reference image and require agents to identify the corresponding item on the website before retrieving information about it or performing an action. We select reference images such that they cannot be identified from textual information alone, requiring models to understand the visual content of the images. Some tasks use images of items that are not available on the website (absent from the store or forum, or already sold on the marketplace), requiring \texttt{N/A} as the correct response. For each task, we provide the agent with two complementary observations: a text representation of the page and a screenshot of the full page. We assign the same index to each link and form in both representations, allowing the agent to identify the corresponding webpage elements across them. We evaluate these tasks using the same framework as the text-based tasks.

\begin{table*}[htb!]
\centering
\small
\setlength{\tabcolsep}{8pt}
\caption{Web agent performance relative to English. The English column reports the accuracy $A_{\text{English}}$, while $\Delta$ Language ($A_{language} - A_{\text{English}}$) denotes the percentage-point difference in accuracy from English for each language $A_{\text{language}}$. Open-source models exhibit substantially larger performance drops in non-English languages, while Gemini-3.1-Flash-lite has the smallest cross-lingual degradation.}
\begin{tabular}{ccccccc}
\toprule
\textbf{Type} & \textbf{Model} & \textbf{EN} & \textbf{$\Delta$ BN} & \textbf{$\Delta$ HI} & \textbf{$\Delta$ UR} & \textbf{$\Delta$ AM} \\
\midrule  

\multirow{3}{*}{\makecell{Closed\\Source}}
& Gemini-3.1-Flash-Lite
& \cellcolor{green!10}{86.40\%}
& \cellcolor{red!10}{-3.70\%}
& \cellcolor{red!10}{-1.50\%}
& \cellcolor{red!10}{-4.00\%}
& \cellcolor{red!10}{-7.30\%} \\

& GPT-5.6 Luna
& \cellcolor{green!10}{85.30\%}
& \cellcolor{red!10}{-7.30\%}
& \cellcolor{red!10}{-6.60\%}
& \cellcolor{red!10}{-9.30\%}
& \cellcolor{red!10}{-16.60\%} \\

& GPT-6 Luna
& \cellcolor{green!10}{83.60\%}
& \cellcolor{red!10}{-11.20\%}
& \cellcolor{red!10}{-8.90\%}
& \cellcolor{red!10}{-15.60\%}
& \cellcolor{red!10}{-14.70\%} \\

\midrule

\multirow{6}{*}{\makecell{Open\\Source}}

& Llama-3.1-8B
& \cellcolor{green!10}{40.90\%}
& \cellcolor{red!10}{-16.00\%}
& \cellcolor{red!10}{-11.60\%}
& \cellcolor{red!10}{-12.70\%}
& \cellcolor{red!10}{-34.50\%} \\

& Gemma3-4B
& \cellcolor{green!10}{12.70\%}
& \cellcolor{red!10}{-1.80\%}
& \cellcolor{green!10}{0.00\%}
& \cellcolor{red!10}{-4.70\%}
& \cellcolor{red!10}{-6.70\%} \\

& Gemma3-12B
& \cellcolor{green!10}{45.10\%}
& \cellcolor{red!10}{-13.10\%}
& \cellcolor{red!10}{-8.70\%}
& \cellcolor{red!10}{-11.80\%}
& \cellcolor{red!10}{-15.80\%} \\

& Gemma3-27B
& \cellcolor{green!10}{52.90\%}
& \cellcolor{red!10}{-9.80\%}
& \cellcolor{red!10}{-12.20\%}
& \cellcolor{red!10}{-11.10\%}
& \cellcolor{red!10}{-18.70\%} \\

& GPT-OSS-20B
& \cellcolor{green!10}{64.20\%}
& \cellcolor{red!10}{-14.40\%}
& \cellcolor{red!10}{-10.40\%}
& \cellcolor{red!10}{-15.10\%}
& \cellcolor{red!10}{-33.30\%} \\

\bottomrule
\end{tabular}
\label{tab:multilingual-delta}
\vspace{-5mm}
\end{table*}

\subsection{Interface Disruptions}

Interface disruptions are common on real-world websites, requiring web agents to navigate them correctly before proceeding with a task \cite{anupam2025browserarena}. To evaluate whether such disruptions affect web agent performance differently across languages, we introduce an optional evaluation setting that includes \textit{cookie notices} and \textit{advertising pop-ups} on the websites. When enabled, each website visit presents two language-localized disruptions in a fixed order. First, a cookie notice must be dismissed using a \textit{close button} before the agent can enter the site. Second, an advertising offer to install an app may take the agent away from the website. Dismissing each disruption reveals the originally requested page, while failing to dismiss either disruption results in task failure. We evaluate agents with and without these disruptions on the same tasks and compare success rates across languages. Further implementation details are provided in Appendix \ref{Appendix:Imple}.

\subsection{Evaluation Protocol}
\label{sec:eval}

A task is considered successful only if all associated evaluation checks pass. We describe the evaluation protocol for each task below, and additional details about each of our separate websites and their implementations in Appendices \ref{Appendix:Imple} and \ref{Appendix: website}.
\vspace{0.5mm}

\noindent\textbf{Information-seeking tasks.}
The agent's final answer is normalized for Unicode, hidden characters, and numeral formats before comparison with the expected answer. An answer is accepted if it contains an expected value and does not contain any explicitly excluded alternatives, preventing ambiguous or hedged responses.
\vspace{0.5mm}

\looseness-1\noindent\textbf{State-changing tasks.}
For state changing tasks, we aim to verify the resulting website's state against the task's ground truth for state-changing tasks. We check the existence, absence, or values of database records, as well as the final page state. Tasks that imply exclusive changes are additionally verified by checking that no unintended state changes occurred.
\vspace{0.5mm}

\noindent\textbf{Unachievable tasks.}
When a task is unachievable, the agent must indicate this by returning \texttt{N/A}. Having described the evaluation protocol, we now present the experimental setup and results in the following sections.
\vspace{0.5mm}

% agent interaction
% episode structure
% success criteria
% state-based checker
% screenshots etc.

\section{Experiments and Results}
In this section, we first describe our experimental setup, and then examine how web agent performance varies across LRLs relative to English. Then, we investigate the impact of interface disruptions and evaluate whether post-training improves agentic performance for LRLs.

\subsection{Experimental Setup}
%\vspace{2mm}

 We evaluate five open-source LLMs: Llama-3.1-8B \cite{grattafiori2024llama}, Gemma3-4B, Gemma3-12B, Gemma3-27B \cite{team2025gemma}, and GPT-OSS-20B \cite{agarwal2025gpt}, spanning 4B to 27B parameters in scale. We also evaluate 3 closed-source models: GPT-6 Luna, GPT-5.6 Luna, and Gemini 3.1 Flash-Lite.

\subsection{(RQ1) Multilingual Performance}

Table \ref{tab:multilingual-delta} summarizes web agent performance across all five languages relative to English. We observe a substantial multilingual performance gap, with the degradation being especially pronounced among open-source models. This is a pertinent issue for practical deployment, where open-source models provide an accessible alternative for practitioners, researchers, and developers compared to the expensive closed-source models.

\looseness-1 Closed-source LLMs generally attain strong performance in English. However, beyond English, performance degrades across other LRLs. Notably, Gemini-3.1-Flash-lite shows the smallest degradation, with performance dropping by only 1.5\% to 7.3\% percentage points across the four low-resource domains. GPT-5.6 and GPT-6 exhibit larger drops; e.g., GPT-6 sees significant degradation for Urdu with a drop of 15.6\% and GPT-5.6 shows a reduction of 16.6\% for Amharic.

Further, open-source LLMs experience severe degradation in non-English domains. Llama-3.1-8B drops by 16.0\% percentage points in Bengali and 34.5\% percentage points in Amharic, while GPT-OSS-20B drops by 14.4\%, 10.4\%, 15.1\%, and 33.3\% percentage points in Bengali, Hindi, Urdu, and Amharic, respectively. Gemma3-12B and Gemma3-27B also exhibit substantial degradation across languages. We can thus infer that while open-source LLMs can still achieve reasonable performance in English (although this is still significantly lower than their closed-source counterparts, as is expected), their performance is considerably worse for agentic tasks in LRLs.
The disparity is more apparent for Amharic, which produces the largest performance drops for most LLMs. 

\begin{figure}[!t]
    \centering
    \includegraphics[width=0.66\columnwidth]{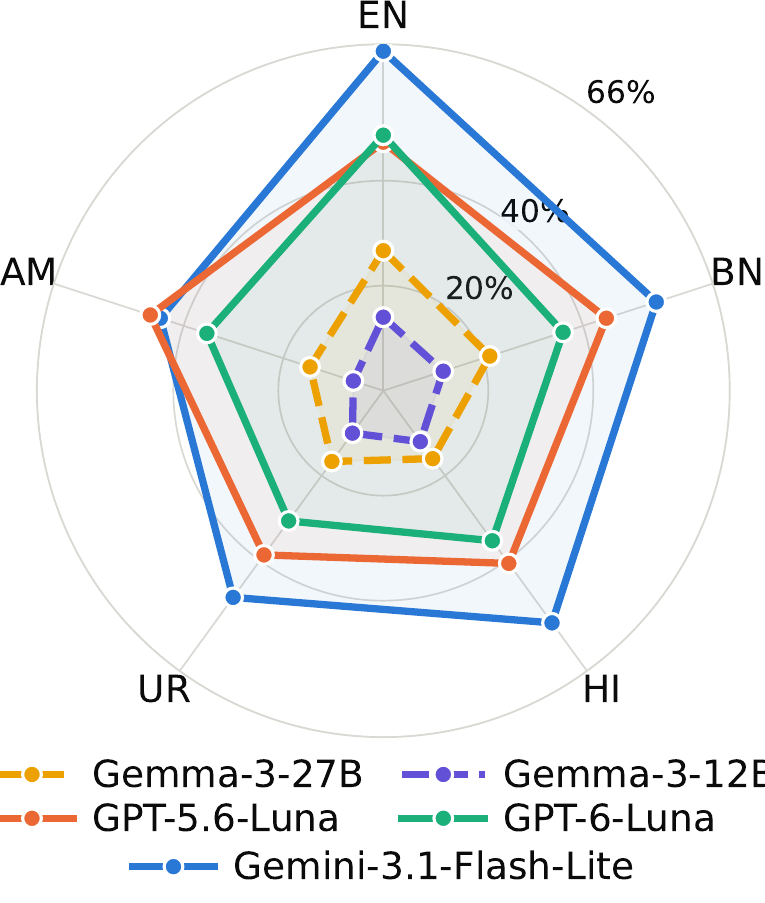}
    \vspace{-3mm}
    \caption{Results on the 150 visual tasks across 3 websites and all 5 languages. Performance of open-source LLMs is significantly lower than the closed-source models (Gemini, GPT-5.6, and GPT-6).}
    \vspace{-5mm}
    \label{fig:visual_results}
\end{figure}

\looseness-1 Figure \ref{fig:visual_results} demonstrates performance on the visual tasks, obtaining similar trends as the text-only tasks. LRLs show lower performance compared to English, and all LLMs exhibit an overall decrease in performance compared to the text-based tasks (e.g., the best Gemini model only attains $\sim$60\% in English for the visual tasks compared to 86.4\% in the English text-only domain). However, closed-source LLMs have lower performance degradation compared to open-source LLMs across languages.  Further results which are specific to each website are also provided in appendix \ref{Appendix:Results}.
\vspace{0.5mm}

\noindent\textbf{Remark.} All in all, these results show that evaluating web agents only in English can substantially overestimate the capabilities of agents in multilingual web environments. The large degradation observed across models highlights the need for multilingual evaluation as a standard when assessing web agent reliability, moving beyond English.

\begin{figure*}[!t]
    \centering
    \includegraphics[width=.98\textwidth]{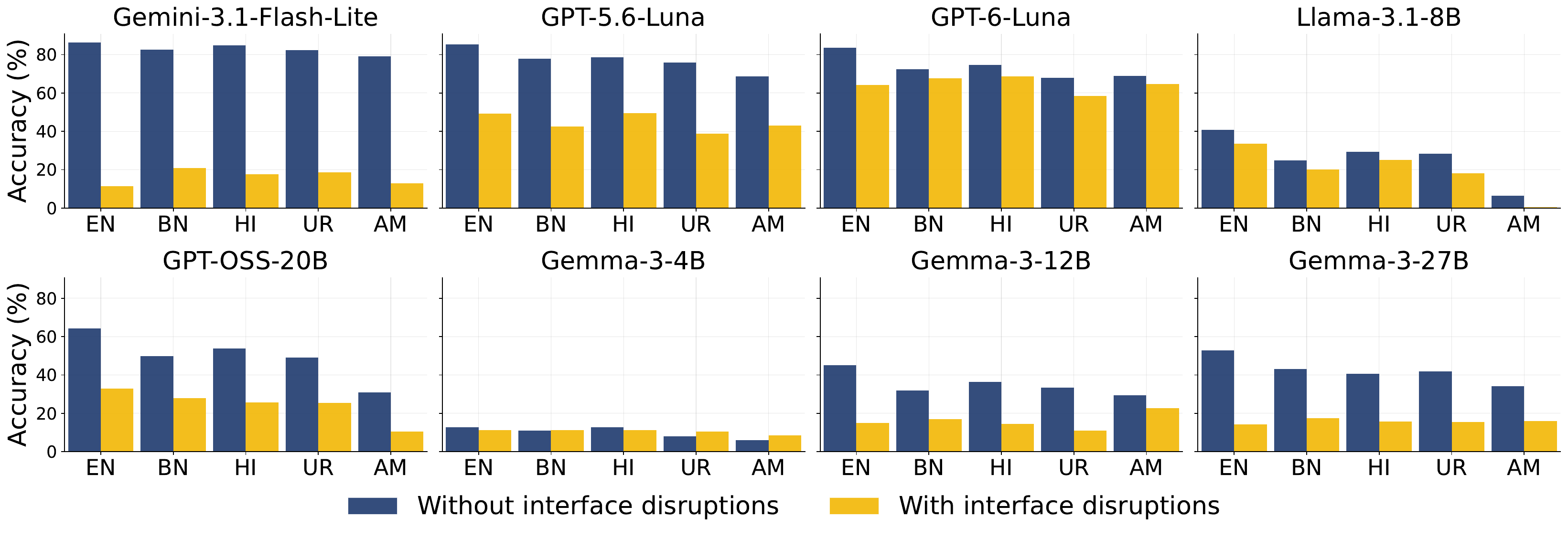}\vspace{-2mm}
    \caption{Performance degradation across LLMs with and without interface disruptions (cookie notices and advertisement pop-ups). The cookie notice appears at the beginning of each task and must be dismissed before the agent can proceed. The advertisement pop-up appears after each click on the website, prompting the agent to install an app. Accepting the offer redirects the agent away from the website and results in task failure, so the agent must correctly dismiss the pop-up to continue. All models see substantial performance drops when interface disruptions are introduced, with Gemini-3.1 Flash-Lite showing the largest accuracy decline.}\vspace{-5mm}
    \label{fig:Popup}
\end{figure*}

\subsection{(RQ2) Interface Disruptions}
Another important finding from our experiments is that a high task success rate does not necessarily correlate with \textit{robustness} to interface disruptions. For this research question, we focus exclusively on text-based tasks, given the poor performance of open-source models and the substantial performance degradation observed in closed-source models on visual tasks. 

\looseness-1 Gemini 3.1 Flash-Lite, which is the highest-performing LLM without interface disruptions, with success rates near 80\% across languages, experiences the largest drop in performance when interface disruptions are introduced (below 20\% across languages). In contrast, GPT-6 Luna decreases comparatively less, becoming the highest-performing model in our experiments with interface disruptions.

Observing results from the Gemma family of models, we can note that scaling to a larger parameter set (i.e. from the 4B-27B range) does not necessarily correlate with robustness in interface disruptions. For both the Gemma3-12B and Gemma3-27B, accuracy dropped by more than half across most languages. Even the strongest open-source model in our experiments, GPT-OSS-20B, halved in performance with the introduction of interface disruptions.\vspace{0.5mm}

\noindent\textbf{Remark}.These results show that seemingly minor interface disruptions can substantially alter both the overall and multilingual performance of web agents. Further, model scale and language resource availability do not directly indicate robustness in downstream agentic performance.

\begin{figure}[!t]
    \centering
    \includegraphics[width=\columnwidth]{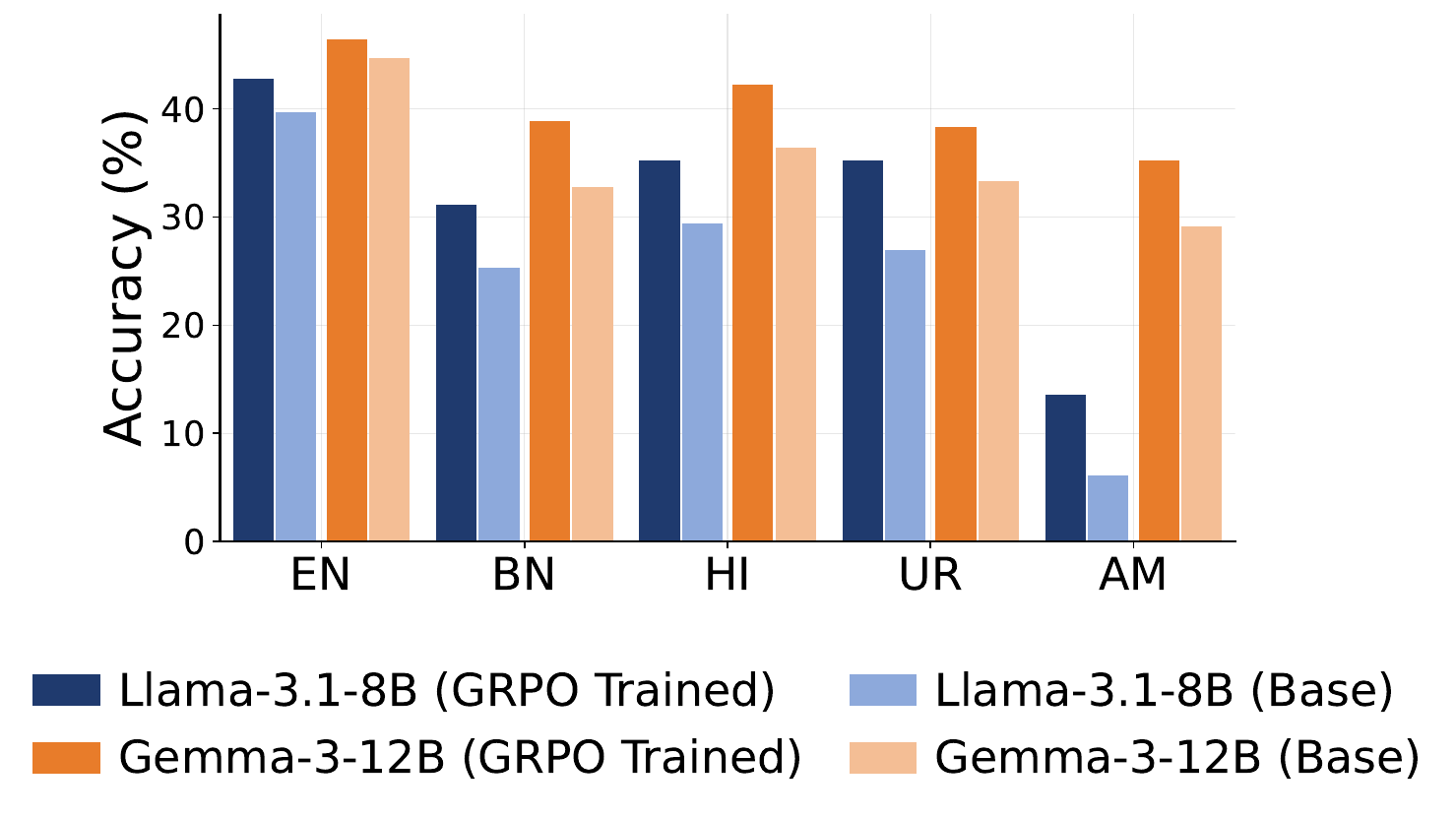}
    \vspace{-8mm}
    \caption{Results on the test subset for Llama-3.1-8B and Gemma3-12B after GRPO training on \textsc{LoWAgent}. Both models show substantial improvement across all languages, reiterating the effectiveness of \textsc{LoWAgent}.}
    \vspace{-7mm}
    \label{fig:GRPO}
\end{figure}

\subsection{(RQ3) Performance Improvement}

To answer whether post-training on a subset of \textsc{LoWAgent} improves the performance of web agents, we train Llama-3.1-8B and Gemma3-12B with Group Relative Policy Optimization (GRPO) \cite{shao2024deepseekmath} on 15 (out of 75) text-based training tasks from each of the 6 websites across all 5 languages (450 tasks) and test on the remaining tasks (60 per website and 1800 with languages and all the websites combined). At each training step, the agent is given a webpage exactly as it would during evaluation and generates the next action. We sample eight candidate actions for the same state and compare each against the corresponding action in the reference solution. An action receives a reward of 1 if it produces the same effect as the reference action, such as opening the same page or submitting the same form value, and 0 otherwise. The model is then updated using its Low-Rank Adaptation (LoRA) \cite{hu2021lora} parameters, encouraging it to generate actions that are more likely to achieve the desired effect.

Figure \ref{fig:GRPO} compares the results between the base model and GRPO-trained models on the 60 tasks per website on \textsc{LoWAgent}. GRPO training improves performance of both models across all languages. Overall, both Llama and Gemma models see around a 5\% increase in performance via post-training across all languages, but the gains are larger for Bengali, Hindi, Urdu, and Amharic (the low-resource languages) than for English.\vspace{0.5mm}

\noindent\textbf{Remark.} These results suggest that the post-training does more than just teach general action selection or familiarity with the website environment, but can also improve language-specific understanding for agents. However, given the sample inefficiency of RL post-training \cite{sun2026improving}, there is a pressing need to address the data scarcity issue with LRLs \cite{ousidhoum2025building}, which we seek to address through this work in part.

\vspace{-2mm}
\section{Conclusion}
\vspace{-2mm}
 \looseness-1 We introduced \textsc{LoWAgent}, a benchmark for evaluating web agents in low-resource language settings. \textsc{LoWAgent} provides a multilingual evaluation across diverse websites, tasks, and interaction settings, enabling a systematic study of how web agent performance is affected in low-resource languages. Our experiments reveal substantial performance gaps between English and low-resource languages, particularly for open-weight models. We further find that interface disruptions such as cookie notices and advertisement pop-ups can substantially degrade agent performance, with performance in the default setting (no interface disruptions) providing a poor indication of robustness. Finally, we show that post-training on a small subset of benchmark tasks substantially improves performance of open-source models, with larger gains observed for LRLs compared to English. This implies that the observed multilingual gap is not solely due to general agent capabilities, but in part from the limited availability of data and training exposure for LRLs. Our findings highlight the need for web-agent benchmarks that move beyond English-centric evaluation and account for the linguistic challenges encountered in real-world web environments.

%\clearpage

% Bibliography entries for the entire Anthology, followed by custom entries
%\bibliography{anthology,custom}
% Custom bibliography entries only
%\clearpage
\bibliography{refs}
\clearpage
%\columnbreak

\appendix
% \onecolumn
\section*{Appendix}

\section{Implementation}
\label{Appendix:Imple}
\vspace{0.5mm}

\noindent\textbf{Website Implementation.}
Each website is implemented as a Flask application backed by SQLite. Each language version runs on a separate server with its own database, initialized from the same underlying records and localized to the target language. Website text and numerical displays are localized, with two exceptions: Amharic websites use Western numerals throughout, and airline websites retain US dollars and Western numerals in every language. The store, marketplace, and rail websites display prices in local currencies, converted from USD using fixed exchange rates.

\begin{figure*}[htb!]
    \centering
    \includegraphics[width=.99\textwidth]{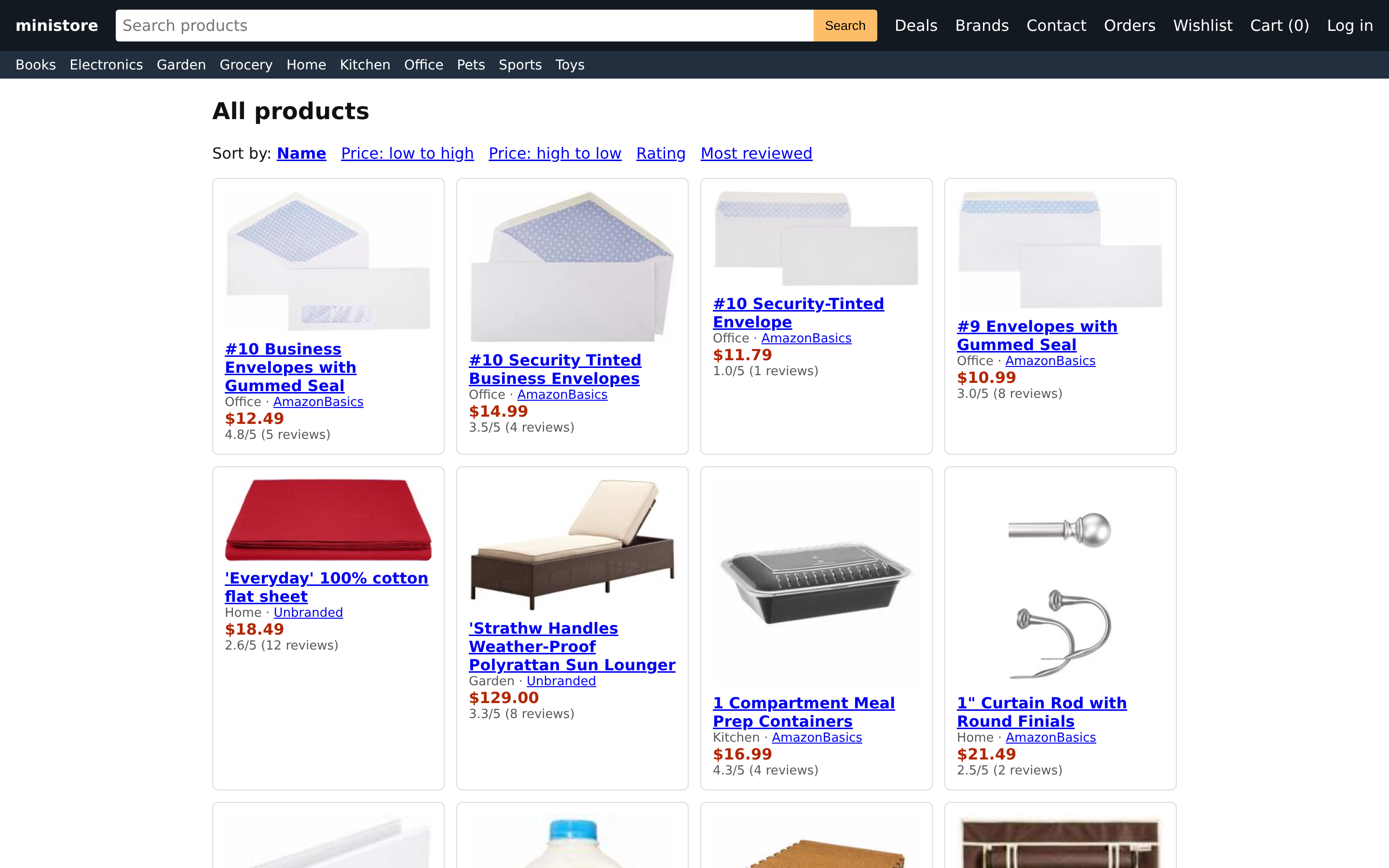}
    \caption{Screenshot of the Ministore website in English, showing its interface and layout as presented to the agent during task execution.}\vspace{-3.5mm}
    \label{fig:ex1}
\end{figure*}

\begin{figure*}[htb!]
    \centering
    \includegraphics[width=.99\textwidth]{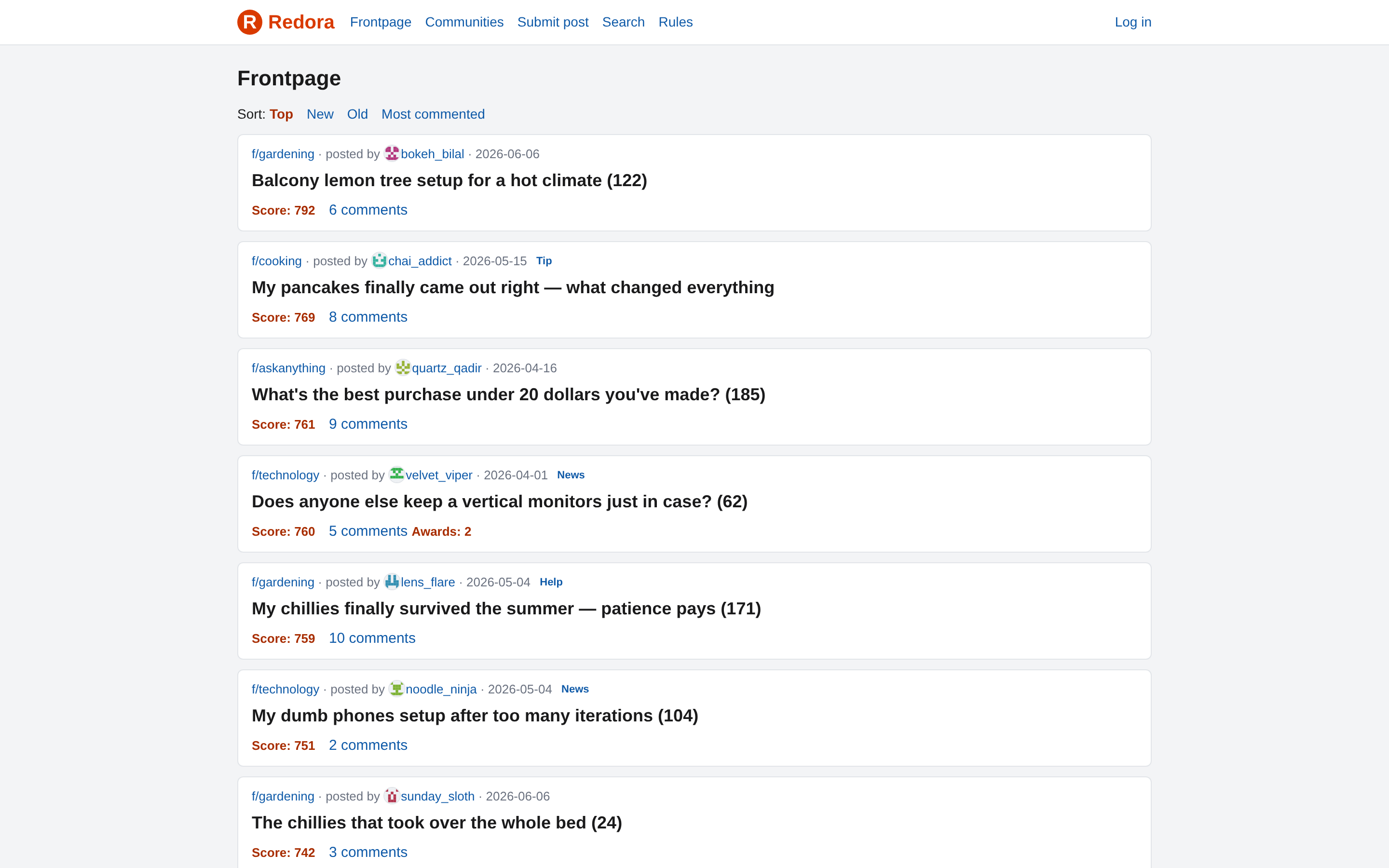}
    \caption{Screenshot of the Redora website in English, showing its interface and layout as presented to the agent during task execution.}\vspace{-3.5mm}
    \label{fig:ex2}
\end{figure*}

\begin{figure*}[htb!]
    \centering
    \includegraphics[width=.99\textwidth]{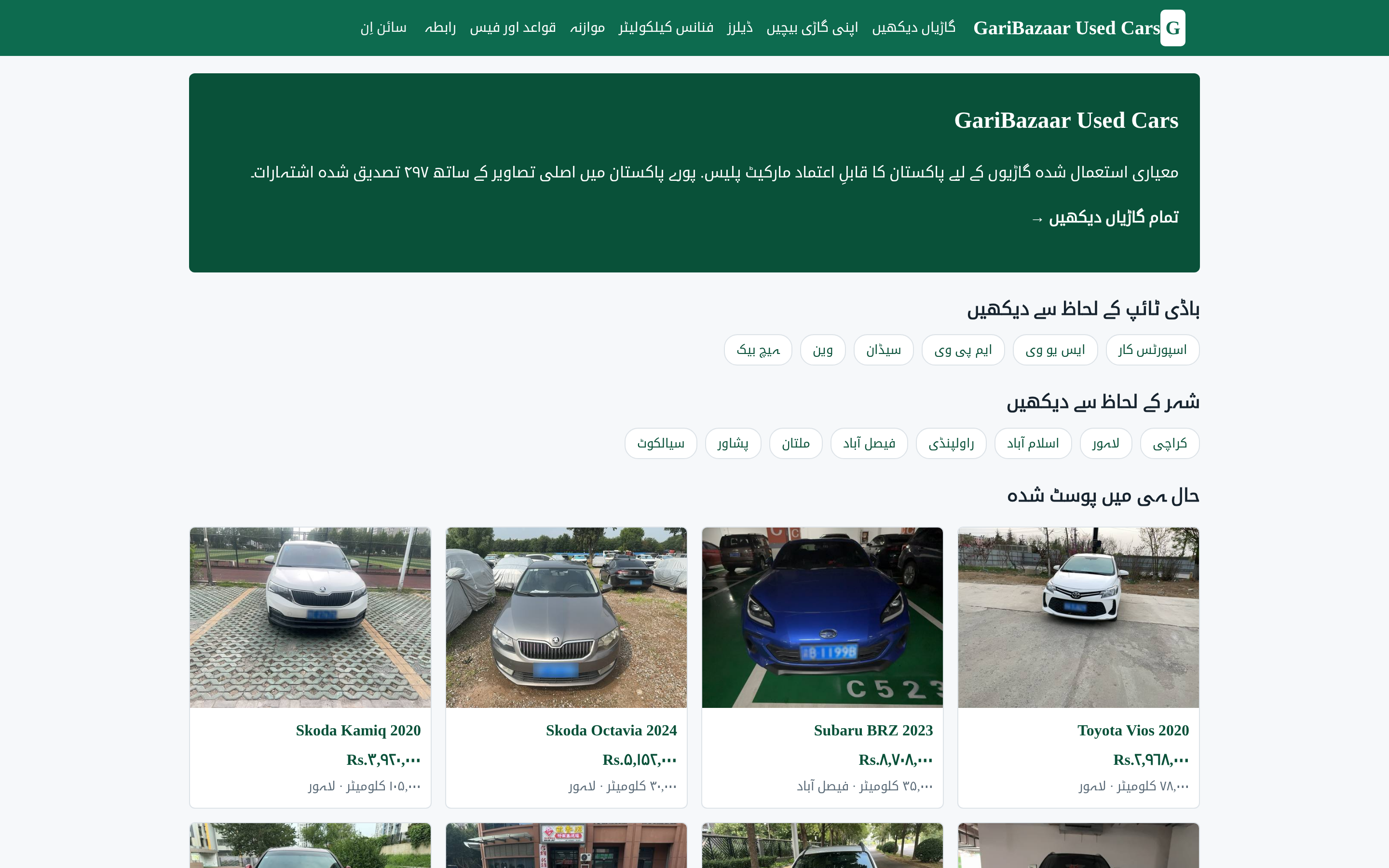}
    \caption{Screenshot of the GariBazaar website in Urdu, showing its interface and layout as presented to the agent during task execution.}\vspace{-3.5mm}
    \label{fig:ex3}
\end{figure*}

\begin{figure*}[htb!]
    \centering
    \includegraphics[width=.99\textwidth]{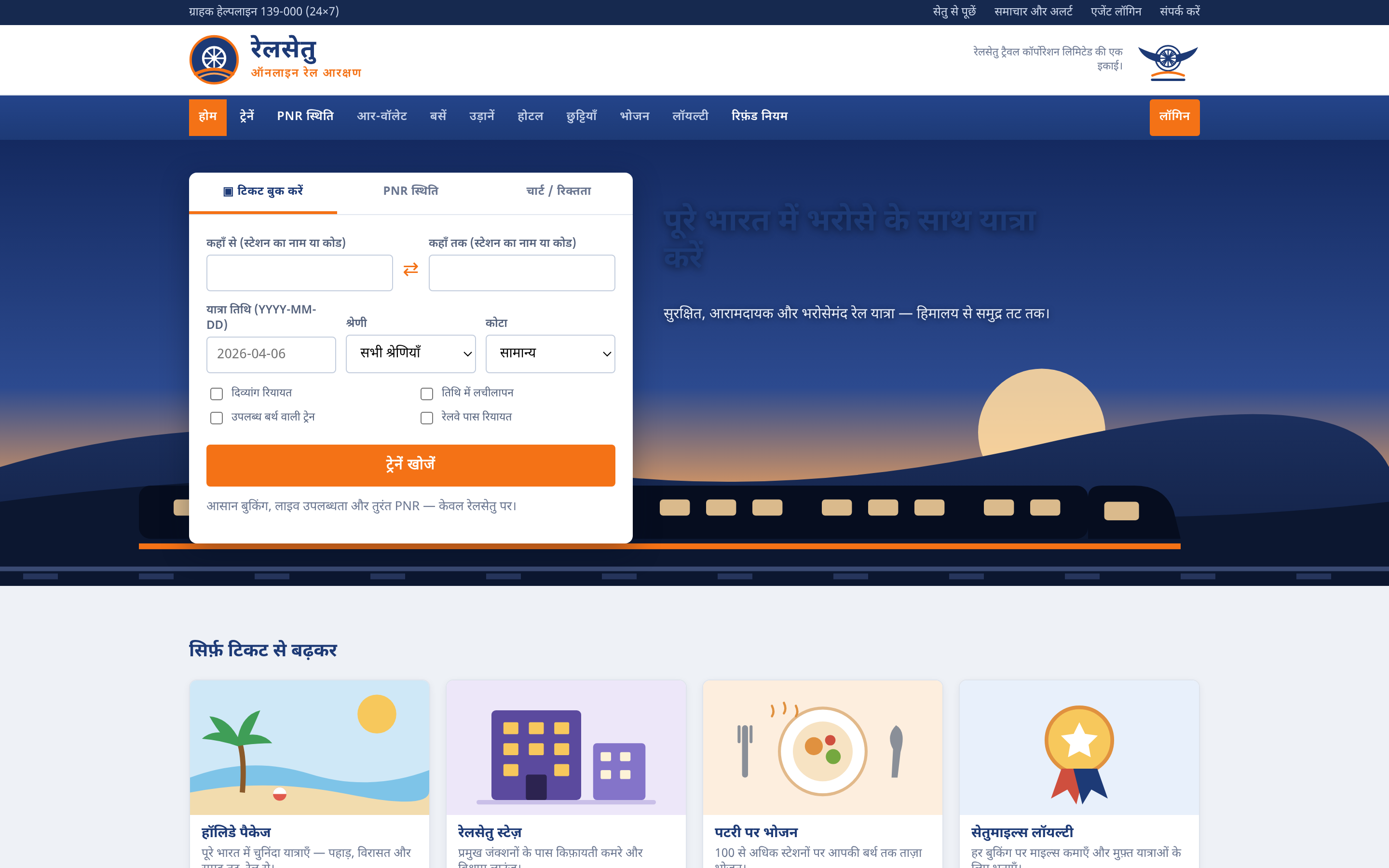}
    \caption{Screenshot of the RailSetu website in Hindi, showing its interface and layout as presented to the agent during task execution.}\vspace{-3.5mm}
    \label{fig:ex4}
\end{figure*}

\begin{figure*}[htb!]
    \centering
    \includegraphics[width=.99\textwidth]{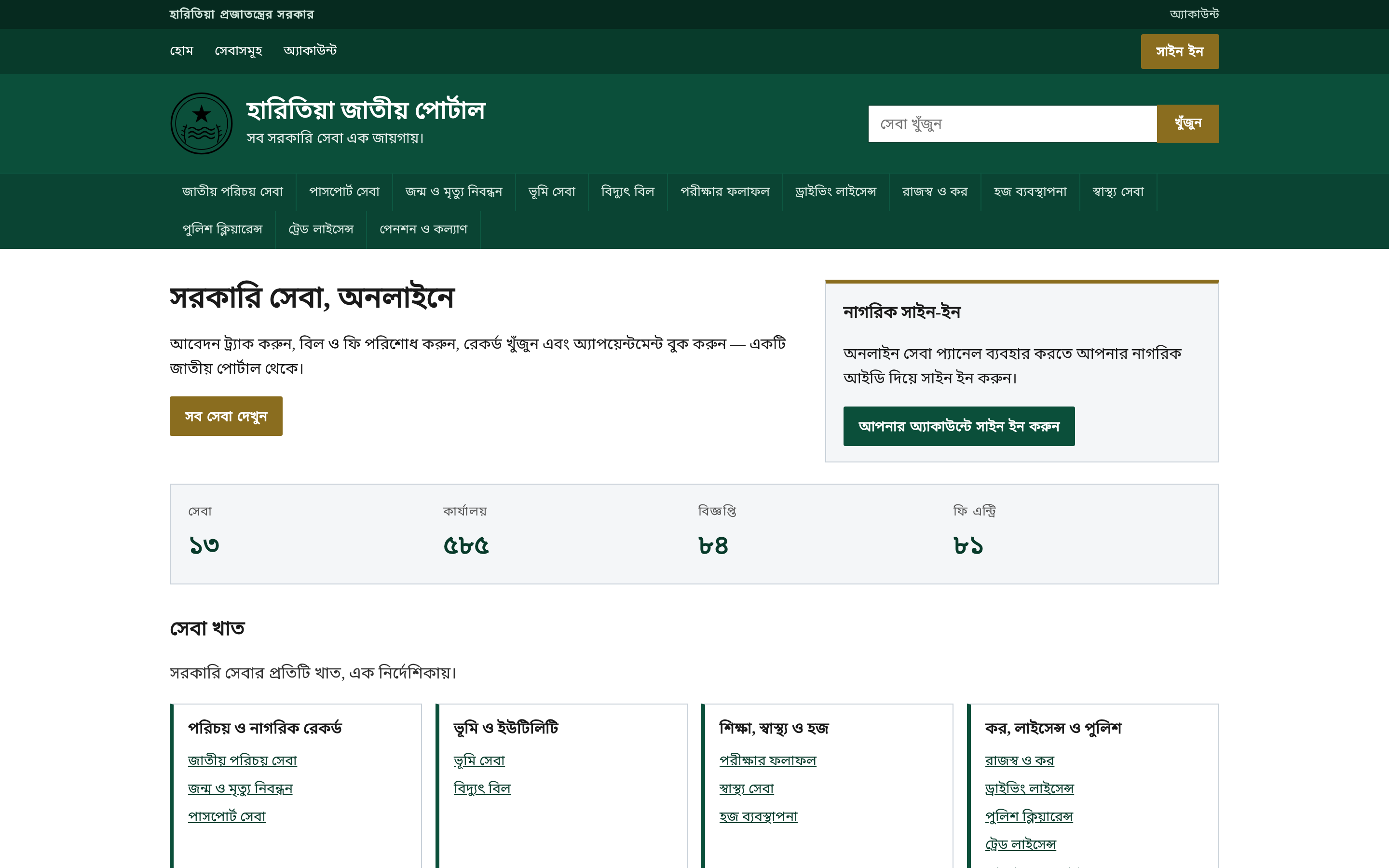}
    \caption{Screenshot of the Haritia National Portal website in Bengali, showing its interface and layout as presented to the agent during task execution.}\vspace{-3.5mm}
    \label{fig:ex5}
\end{figure*}

\begin{figure*}[htb!]
    \centering
    \includegraphics[width=.99\textwidth]{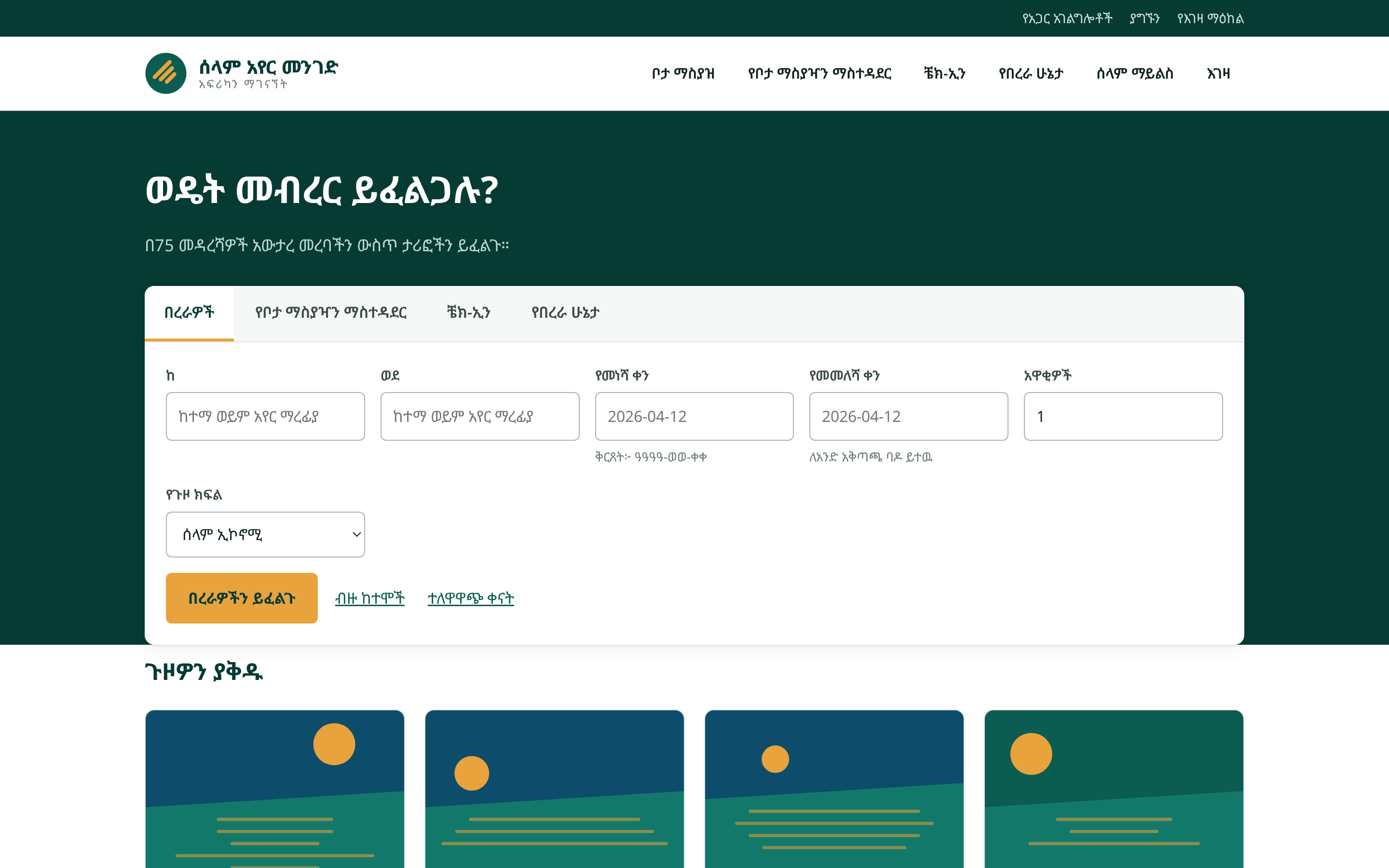}
    \caption{Screenshot of the Selam Airways website in Amharic, showing its interface and layout as presented to the agent during task execution.}\vspace{-3.5mm}
    \label{fig:ex6}
\end{figure*}

Before each episode, the harness restores every website's database to its initial state, ensuring that episodes run independently and can be reproduced. Database resets and state inspection are handled through endpoints protected by a harness-only token, which is inaccessible to the agent. Each website also provides a test account, with credentials included in the task intent, along with any pre-existing user data needed to complete the task, such as past orders, bookings, posts, or saved listings. Screenshots of the website homepages are given in \cref{fig:ex1,fig:ex2,fig:ex3,fig:ex4,fig:ex5,fig:ex6}.

\vspace{0.5mm}
\noindent\textbf{Interface Disruption implementation.}
Interface disruptions (cookie notices and advertisement pop-ups) are built into the websites themselves and appear as ordinary pages that agents must handle using the same browser and action space as the rest of the task. When disruptions are enabled, each episode presents two interruptions in a fixed order. First, a cookie notice appears on arrival and can only be dismissed by clicking its close link. On the next page load, the agent encounters an advertisement pop-up prompting it to install a partner app. Following a common dark-pattern design, the ``Install the app'' link always appears first. Clicking it takes the agent away from the website and ends the episode in failure, while clicking ``No thanks'' returns the agent to the page it originally requested.

\begin{figure*}[htb!]
    \centering
    \includegraphics[width=.99\textwidth]{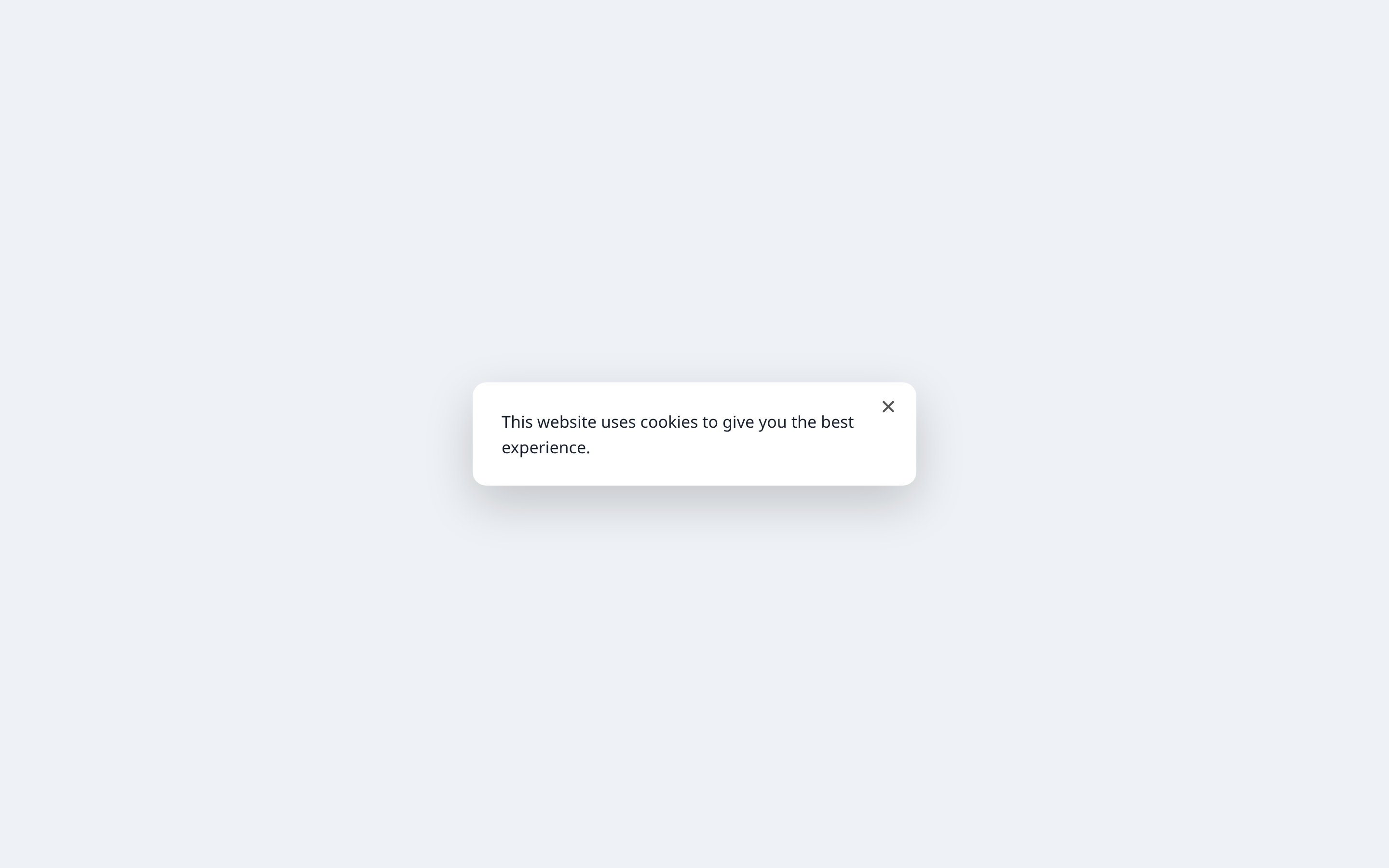}
    \caption{Screenshot of the cookie notice visible to the agent during task execution.}\vspace{-3.5mm}
    \label{fig:cookie}
\end{figure*}

\begin{figure*}[htb!]
    \centering
    \includegraphics[width=.99\textwidth]{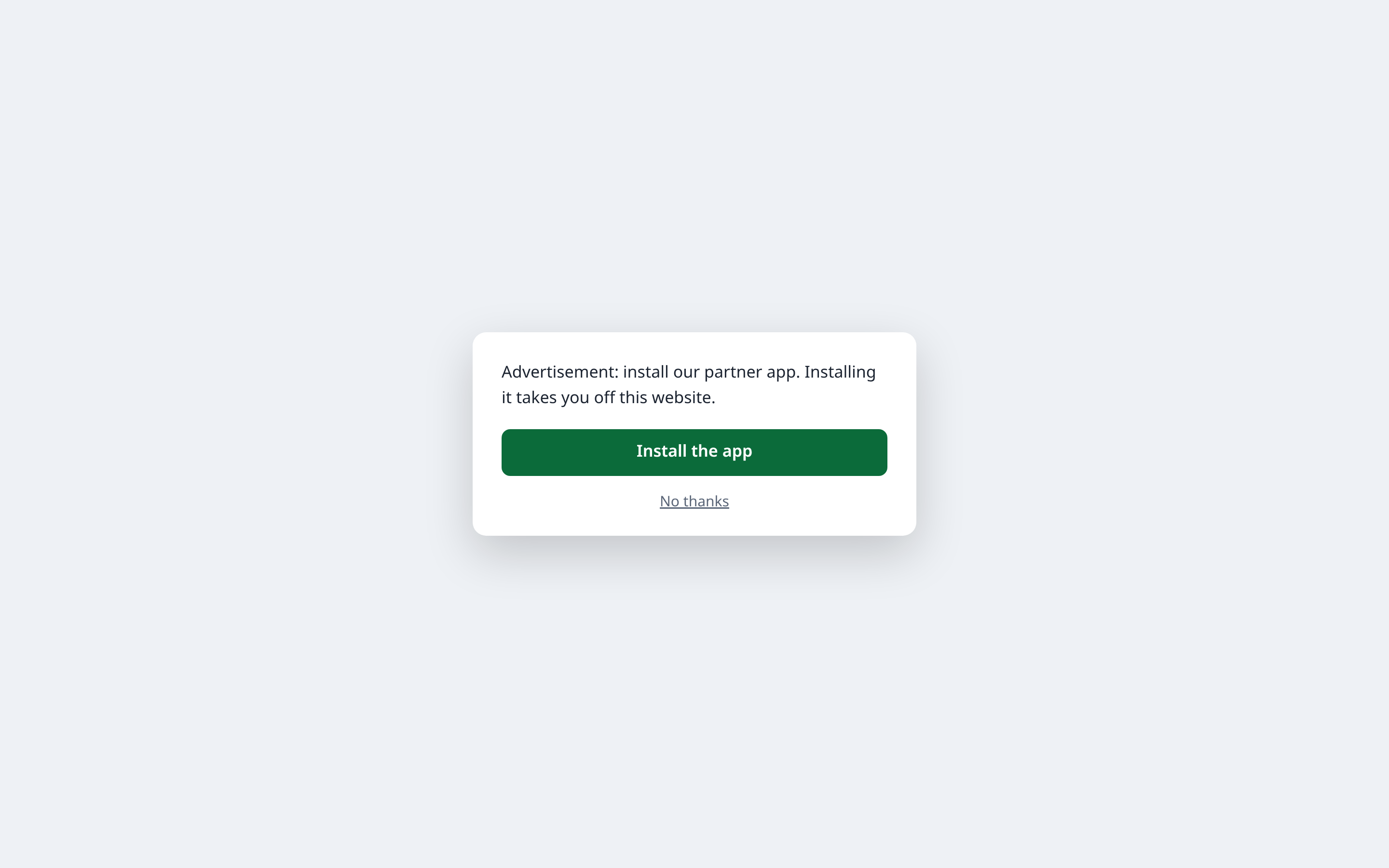}
    \caption{Screenshot of the advertisement pop-up visible to the agent during task execution.}\vspace{-3.5mm}
    \label{fig:ad}
\end{figure*}

While an interruption is active, every page request returns the same disruption, and its links are assigned fresh random URLs on each visit. This prevents agents from bypassing the interruptions through direct navigation or reusing links across episodes. The disruption text is translated into all five languages and verified by native speakers. We also increase each task's step limit by two to account for the additional interactions. For each episode, we record how the agent handles each disruption: whether it dismisses it, accepts the offer, ends the task while the disruption is active, or reaches the step limit. Screenshots of the interface disruptions are presented in \cref{fig:cookie,fig:ad}.

\noindent\textbf{GRPO implementation.}
We implement GRPO with the verl library. Rather than updating all weights, we train Low-Rank Adaptation (LoRA) adapters of rank 32 ($\alpha=64$) on all attention and MLP projection matrices. Each update samples 64 training states and 8 responses per state at temperature 1.0, with responses capped at 256 tokens, enough for one action, and prompts up to 32,768 tokens. We use a learning rate of $10^{-5}$, a KL penalty of 0.001 toward the frozen base model (low-variance KL estimator, applied as a loss term rather than in the reward). Training runs for three epochs (87 updates) on two NVIDIA B200 GPUs in bfloat16 with gradient checkpointing. For evaluation, the LoRA weights are merged into the base model, which is then served with vLLM in the same harness as all other models.

For RQ3, we instantiate our GRPO-based approach in the text-only setting. This already demonstrates that performance can be improved with \textsc{LoWAgent}. We restrict attention to text because extending GRPO to visual inputs is non-trivial: standard GRPO applied to multimodal RL suffers from gradient stagnation that degrades training stability~\cite{chen2025advancing}. Our method is not specific to text, however, and is in principle applicable to the multimodal settings.

\section{Inference}
We infer all open-weight models locally using vLLM on NVIDIA B200 GPUs with BF16 precision (MXFP4 for GPT-OSS), with a maximum context length of 128k tokens. Each open-source model is evaluated using its lowest available reasoning setting: no explicit reasoning for Llama and Gemma, and \emph{low} reasoning effort for GPT-OSS. Greedy decoding ($T=0$) is used to ensure deterministic generation across models. Both open and closed-source models use a maximum of 4,096 output tokens per step. Similar to open-source models, closed-source models in our experiments are evaluated using their lowest available reasoning setting: a thinking budget of 0 for Gemini and \emph{none} reasoning effort for both GPT models. We use each provider's default sampling parameters. All models use the same agent configuration, including a shared system prompt specifying the action syntax, one action per response, and test-account credentials.

\section{Websites}
\label{Appendix: website}
\vspace{0.5mm}
\noindent\textbf{Ministore} is an e-commerce store with 2,565 products across 10 categories, shown with catalogue photos, alongside 12,279 reviews, 1,108 Q\&A answers, and 99 brands. Users can search for items, go through a multi-step checkout, apply coupons, and manage a wishlist. They can also edit or cancel orders, leave reviews, ask and answer questions, and mark responses as helpful. We source the product images used in Ministore from the ABO dataset \cite{collins2022abo}.

\vspace{0.5mm}
\noindent\textbf{Redora} is a discussion forum built around 12 communities, with 2,000 posts (500 of them with photos) and 12,094 comments. People can vote, comment and reply, create posts, and edit or delete their own. They can also subscribe to communities, save posts, report content, and browse profiles and community pages. Redora contains images from both the ABO \cite{collins2022abo} and 3DRealCar dataset \cite{du20253drealcar}.

\vspace{0.5mm}
\noindent\textbf{GariBazaar} is a used-car marketplace with 497 listings, each carrying six photos, drawn from 10 dealers and paired with inspection reports and price-drop history. Buyers can run filtered searches, make offers subject to minimum-price rules, save cars, and compare options. They can also post their own ads and work out payments with a finance calculator. The car images are taken from the 3DRealCar dataset \cite{du20253drealcar}.

\vspace{0.5mm}
\noindent\textbf{RailSetu} is a railway ticketing site covering 22 Indian stations and 13 trains, with full timetables and fares. Users can search for trains, check schedules and seat availability, and book tickets with a PNR. They can also look up PNR status and handle cancellations and refunds.

\vspace{0.5mm}
\noindent\textbf{Haritia National Portal} is a government services site offering 13 services, each with 10 sub-services, spanning identity, passport, land, tax, utilities, exams, and civil registration. Users can submit applications, book appointments, and pay bills and taxes. They can also look up records and check fee schedules.

\vspace{0.5mm}
\noindent\textbf{Selam Airways} is an airline site handling 133 routes, with bookings, a loyalty programme, lounges, and partner services. Travellers can manage bookings, choose seats, add baggage and meals, and request upgrades. They can also process refunds and cancellations, check in, manage loyalty accounts, and review fare rules.

\section{Additional Results}
\label{Appendix:Results}
This section reports per-website results that are aggregated in the main paper. For every model, we break down task success by website and language, both on the standard benchmark (RQ1) shown in \cref{tab:gemini-site-language,tab:gpt56-site-language,tab:gpt6-site-language,tab:llama32-site-language,tab:llama31-site-language,tab:gemma34b-site-language,tab:gemma312b-site-language,tab:gemma327b-site-language,tab:gptoss20b-site-language,tab:appendix-multilingual-results} and with interface disruptions enabled (RQ2) shown in \cref{tab:gemini-popups-site-language,tab:gpt56luna-popups-site-language,tab:gpt6luna-popups-site-language,tab:llama8b-popups-site-language,tab:gemma4b-popups-site-language,tab:gemma12b-popups-site-language,tab:gemma27b-popups-site-language,tab:gptoss20b-popups-site-language}. Each cell gives the success rate over the 75 tasks of one website in one language.

\begin{table*}[t]
\centering
\small
\setlength{\tabcolsep}{12pt}
\caption{Gemini-3.1-Flash-Lite success rates across the six websites and five languages.}
\begin{tabular}{lccccc c}
\toprule
\textbf{Website} & \textbf{EN} & \textbf{BN} & \textbf{HI} & \textbf{UR} & \textbf{AM} & \textbf{Overall} \\
\midrule
Ministore & 93.3\% & 90.7\% & 88.0\% & 92.0\% & 86.7\% & 90.1\% \\
Redora & 85.3\% & 85.3\% & 84.0\% & 81.3\% & 82.7\% & 83.7\% \\
GariBazaar & 85.3\% & 85.3\% & 85.3\% & 84.0\% & 72.0\% & 82.4\% \\
Haritia Portal & 76.0\% & 69.3\% & 80.0\% & 70.7\% & 73.3\% & 73.9\% \\
RailSetu & 82.7\% & 76.0\% & 81.3\% & 78.7\% & 76.0\% & 78.9\% \\
Selam Airways & 96.0\% & 89.3\% & 90.7\% & 88.0\% & 84.0\% & 89.6\% \\
\bottomrule
\end{tabular}
\label{tab:gemini-site-language}
\end{table*}

\begin{table*}[t]
\centering
\small
\setlength{\tabcolsep}{12pt}
\caption{GPT-5.6 Luna success rates across the six websites and five languages.}
\begin{tabular}{lccccc c}
\toprule
\textbf{Website} & \textbf{EN} & \textbf{BN} & \textbf{HI} & \textbf{UR} & \textbf{AM} & \textbf{Overall} \\
\midrule
Ministore & 93.3\% & 72.0\% & 73.3\% & 61.3\% & 57.3\% & 71.5\% \\
Redora & 80.0\% & 77.3\% & 78.7\% & 78.7\% & 77.3\% & 78.4\% \\
GariBazaar & 86.7\% & 84.0\% & 78.7\% & 80.0\% & 68.0\% & 79.5\% \\
Haritia Portal & 77.3\% & 70.7\% & 74.7\% & 77.3\% & 64.0\% & 72.8\% \\
RailSetu & 81.3\% & 76.0\% & 80.0\% & 77.3\% & 64.0\% & 75.7\% \\
Selam Airways & 93.3\% & 88.0\% & 86.7\% & 81.3\% & 81.3\% & 86.1\% \\
\bottomrule
\end{tabular}
\label{tab:gpt56-site-language}
\end{table*}

\begin{table*}[t]
\centering
\small
\setlength{\tabcolsep}{12pt}
\caption{GPT-6 Luna success rates across the six websites and five languages.}
\begin{tabular}{lccccc c}
\toprule
\textbf{Website} & \textbf{EN} & \textbf{BN} & \textbf{HI} & \textbf{UR} & \textbf{AM} & \textbf{Overall} \\
\midrule
Ministore & 90.7\% & 64.0\% & 61.3\% & 44.0\% & 48.0\% & 61.6\% \\
Redora & 86.7\% & 76.0\% & 78.7\% & 86.7\% & 78.7\% & 81.3\% \\
GariBazaar & 85.3\% & 82.7\% & 78.7\% & 80.0\% & 73.3\% & 80.0\% \\
Haritia Portal & 76.0\% & 76.0\% & 74.7\% & 70.7\% & 68.0\% & 73.1\% \\
RailSetu & 73.3\% & 62.7\% & 72.0\% & 65.3\% & 64.0\% & 67.5\% \\
Selam Airways & 89.3\% & 73.3\% & 82.7\% & 61.3\% & 81.3\% & 77.6\% \\
\bottomrule
\end{tabular}
\label{tab:gpt6-site-language}
\end{table*}

\begin{table*}[t]
\centering
\small
\setlength{\tabcolsep}{12pt}
\caption{Llama-3.2-3B success rates across the six websites and five languages.}
\begin{tabular}{lccccc c}
\toprule
\textbf{Website} & \textbf{EN} & \textbf{BN} & \textbf{HI} & \textbf{UR} & \textbf{AM} & \textbf{Overall} \\
\midrule
Ministore & 5.3\% & 0.0\% & 6.7\% & 4.0\% & 0.0\% & 3.2\% \\
Redora & 2.7\% & 1.3\% & 1.3\% & 0.0\% & 0.0\% & 1.1\% \\
GariBazaar & 2.7\% & 1.3\% & 0.0\% & 1.3\% & 0.0\% & 1.1\% \\
Haritia Portal & 5.3\% & 6.7\% & 2.7\% & 0.0\% & 0.0\% & 2.9\% \\
RailSetu & 9.3\% & 5.3\% & 8.0\% & 4.0\% & 0.0\% & 5.3\% \\
Selam Airways & 4.0\% & 0.0\% & 2.7\% & 1.3\% & 0.0\% & 1.6\% \\
\bottomrule
\end{tabular}
\label{tab:llama32-site-language}
\end{table*}

\begin{table*}[t]
\centering
\small
\setlength{\tabcolsep}{12pt}
\caption{Llama-3.1-8B success rates across the six websites and five languages.}
\begin{tabular}{lccccc c}
\toprule
\textbf{Website} & \textbf{EN} & \textbf{BN} & \textbf{HI} & \textbf{UR} & \textbf{AM} & \textbf{Overall} \\
\midrule
Ministore & 28.0\% & 22.7\% & 25.3\% & 26.7\% & 6.7\% & 21.9\% \\
Redora & 37.3\% & 29.3\% & 29.3\% & 30.7\% & 1.3\% & 25.6\% \\
GariBazaar & 36.0\% & 17.3\% & 26.7\% & 21.3\% & 4.0\% & 21.1\% \\
Haritia Portal & 42.7\% & 28.0\% & 28.0\% & 37.3\% & 16.0\% & 30.4\% \\
RailSetu & 46.7\% & 28.0\% & 29.3\% & 28.0\% & 5.3\% & 27.5\% \\
Selam Airways & 54.7\% & 24.0\% & 37.3\% & 25.3\% & 5.3\% & 29.3\% \\
\bottomrule
\end{tabular}
\label{tab:llama31-site-language}
\end{table*}

\begin{table*}[t]
\centering
\small
\setlength{\tabcolsep}{12pt}
\caption{Gemma3-4B success rates across the six websites and five languages.}
\begin{tabular}{lccccc c}
\toprule
\textbf{Website} & \textbf{EN} & \textbf{BN} & \textbf{HI} & \textbf{UR} & \textbf{AM} & \textbf{Overall} \\
\midrule
Ministore & 9.3\% & 5.3\% & 8.0\% & 8.0\% & 9.3\% & 8.0\% \\
Redora & 8.0\% & 8.0\% & 9.3\% & 9.3\% & 5.3\% & 8.0\% \\
GariBazaar & 14.7\% & 14.7\% & 12.0\% & 10.7\% & 8.0\% & 12.0\% \\
Haritia Portal & 6.7\% & 8.0\% & 12.0\% & 2.7\% & 4.0\% & 6.7\% \\
RailSetu & 25.3\% & 17.3\% & 21.3\% & 14.7\% & 6.7\% & 17.1\% \\
Selam Airways & 12.0\% & 12.0\% & 13.3\% & 2.7\% & 2.7\% & 8.5\% \\
\bottomrule
\end{tabular}
\label{tab:gemma34b-site-language}
\end{table*}

\begin{table*}[t]
\centering
\small
\setlength{\tabcolsep}{12pt}
\caption{Gemma3-12B success rates across the six websites and five languages.}
\begin{tabular}{lccccc c}
\toprule
\textbf{Website} & \textbf{EN} & \textbf{BN} & \textbf{HI} & \textbf{UR} & \textbf{AM} & \textbf{Overall} \\
\midrule
Ministore & 37.3\% & 25.3\% & 33.3\% & 22.7\% & 24.0\% & 28.5\% \\
Redora & 38.7\% & 29.3\% & 37.3\% & 38.7\% & 34.7\% & 35.7\% \\
GariBazaar & 48.0\% & 29.3\% & 30.7\% & 26.7\% & 22.7\% & 31.5\% \\
Haritia Portal & 38.7\% & 38.7\% & 34.7\% & 36.0\% & 24.0\% & 34.4\% \\
RailSetu & 53.3\% & 34.7\% & 41.3\% & 40.0\% & 37.3\% & 41.3\% \\
Selam Airways & 54.7\% & 34.7\% & 41.3\% & 36.0\% & 33.3\% & 40.0\% \\
\bottomrule
\end{tabular}
\label{tab:gemma312b-site-language}
\end{table*}

\begin{table*}[t]
\centering
\small
\setlength{\tabcolsep}{12pt}
\caption{Gemma3-27B success rates for each website and language.}
\begin{tabular}{lccccc c}
\toprule
\textbf{Website} & \textbf{EN} & \textbf{BN} & \textbf{HI} & \textbf{UR} & \textbf{AM} & \textbf{Overall} \\
\midrule
Ministore & 46.7\% & 40.0\% & 30.7\% & 28.0\% & 26.7\% & 34.4\% \\
Redora & 46.7\% & 34.7\% & 28.0\% & 34.7\% & 28.0\% & 34.4\% \\
GariBazaar & 50.7\% & 45.3\% & 40.0\% & 44.0\% & 30.7\% & 42.1\% \\
Haritia Portal & 46.7\% & 36.0\% & 34.7\% & 37.3\% & 28.0\% & 36.5\% \\
RailSetu & 62.7\% & 50.7\% & 52.0\% & 52.0\% & 44.0\% & 52.3\% \\
Selam Airways & 64.0\% & 52.0\% & 58.7\% & 54.7\% & 48.0\% & 55.5\% \\
\bottomrule
\end{tabular}
\label{tab:gemma327b-site-language}
\end{table*}

\begin{table*}[t]
\centering
\small
\setlength{\tabcolsep}{12pt}
\caption{GPT-OSS-20B success rates for each website and language.}
\begin{tabular}{lccccc c}
\toprule
\textbf{Website} & \textbf{EN} & \textbf{BN} & \textbf{HI} & \textbf{UR} & \textbf{AM} & \textbf{Overall} \\
\midrule
Ministore & 77.3\% & 38.7\% & 45.3\% & 37.3\% & 26.7\% & 45.1\% \\
Redora & 56.0\% & 53.3\% & 54.7\% & 53.3\% & 36.0\% & 50.7\% \\
GariBazaar & 70.7\% & 49.3\% & 53.3\% & 48.0\% & 24.0\% & 49.1\% \\
Haritia Portal & 41.3\% & 38.7\% & 45.3\% & 46.7\% & 26.7\% & 39.7\% \\
RailSetu & 69.3\% & 56.0\% & 57.3\% & 54.7\% & 38.7\% & 55.2\% \\
Selam Airways & 70.7\% & 62.7\% & 66.7\% & 54.7\% & 33.3\% & 57.6\% \\
\bottomrule
\end{tabular}
\label{tab:gptoss20b-site-language}
\end{table*}

\begin{table*}[t]
\centering
\small
\setlength{\tabcolsep}{12pt}
\caption{Web agent success rates aggregated across all six websites for each language, for closed-source and open-source models.}
\begin{tabular}{llccccc}
\toprule
\textbf{Access} & \textbf{Model} & \textbf{EN} & \textbf{BN} & \textbf{HI} & \textbf{UR} & \textbf{AM} \\
\midrule
\multirow{3}{*}{\makecell{Closed\\Source}} 
& Gemini-3.1-Flash-lite & 86.4\% & 82.7\% & 84.9\% & 82.4\% & 79.1\% \\
& GPT-5.6 Luna & 85.3\% & 78.0\% & 78.7\% & 76.0\% & 68.7\% \\
& GPT-6 Luna & 83.6\% & 72.4\% & 74.7\% & 68.0\% & 68.9\% \\
\midrule
\multirow{6}{*}{\makecell{Open\\Source}}
& Llama-3.2-3B & 4.9\% & 2.4\% & 3.6\% & 1.8\% & 0.0\% \\
& Llama-3.1-8B & 40.9\% & 24.9\% & 29.3\% & 28.2\% & 6.4\% \\
& Gemma3-4B & 12.7\% & 10.9\% & 12.7\% & 8.0\% & 6.0\% \\
& Gemma3-12B & 45.1\% & 32.0\% & 36.4\% & 33.3\% & 29.3\% \\
& Gemma3-27B & 52.9\% & 43.1\% & 40.7\% & 41.8\% & 34.2\% \\
& GPT-OSS-20B & 64.2\% & 49.8\% & 53.8\% & 49.1\% & 30.9\% \\
\bottomrule
\end{tabular}

\label{tab:appendix-multilingual-results}
\end{table*}

\begin{table*}[t]
\centering
\small
\setlength{\tabcolsep}{12pt}
\caption{Gemini-3.1-Flash-Lite success rates with interface disruptions for each website and language.}
\begin{tabular}{lccccc c}
\toprule
\textbf{Website} & \textbf{EN} & \textbf{BN} & \textbf{HI} & \textbf{UR} & \textbf{AM} & \textbf{Overall} \\
\midrule
Ministore & 12.0\% & 17.3\% & 13.3\% & 16.0\% & 12.0\% & 14.1\% \\
Redora & 10.7\% & 14.7\% & 8.0\% & 18.7\% & 6.7\% & 11.7\% \\
GariBazaar & 9.3\% & 26.7\% & 20.0\% & 22.7\% & 10.7\% & 17.9\% \\
Haritia Portal & 5.3\% & 13.3\% & 12.0\% & 8.0\% & 8.0\% & 9.3\% \\
RailSetu & 2.7\% & 13.3\% & 12.0\% & 13.3\% & 5.3\% & 9.3\% \\
Selam Airways & 28.0\% & 40.0\% & 40.0\% & 33.3\% & 34.7\% & 35.2\% \\
\bottomrule
\end{tabular}
\label{tab:gemini-popups-site-language}
\end{table*}

% GPT-5.6-Luna: all six websites 1005/2250 = 44.7%
\begin{table*}[t]
\centering
\small
\setlength{\tabcolsep}{12pt}
\caption{GPT-5.6-Luna success rates with interface disruptions for each website and language.}
\begin{tabular}{lccccc c}
\toprule
\textbf{Website} & \textbf{EN} & \textbf{BN} & \textbf{HI} & \textbf{UR} & \textbf{AM} & \textbf{Overall} \\
\midrule
Ministore & 50.7\% & 40.0\% & 49.3\% & 30.7\% & 26.7\% & 39.5\% \\
Redora & 40.0\% & 42.7\% & 38.7\% & 34.7\% & 41.3\% & 39.5\% \\
GariBazaar & 60.0\% & 53.3\% & 57.3\% & 42.7\% & 42.7\% & 51.2\% \\
Haritia Portal & 49.3\% & 38.7\% & 48.0\% & 30.7\% & 48.0\% & 42.9\% \\
RailSetu & 53.3\% & 42.7\% & 60.0\% & 54.7\% & 49.3\% & 52.0\% \\
Selam Airways & 42.7\% & 37.3\% & 44.0\% & 40.0\% & 50.7\% & 42.9\% \\
\bottomrule
\end{tabular}
\label{tab:gpt56luna-popups-site-language}
\end{table*}

% GPT-6-Luna: all six websites 1456/2250 = 64.7%
\begin{table*}[t]
\centering
\small
\setlength{\tabcolsep}{12pt}
\caption{GPT-6-Luna success rates with interface disruptions for each website and language.}
\begin{tabular}{lccccc c}
\toprule
\textbf{Website} & \textbf{EN} & \textbf{BN} & \textbf{HI} & \textbf{UR} & \textbf{AM} & \textbf{Overall} \\
\midrule
Ministore & 56.0\% & 56.0\% & 58.7\% & 38.7\% & 45.3\% & 50.9\% \\
Redora & 66.7\% & 72.0\% & 72.0\% & 68.0\% & 72.0\% & 70.1\% \\
GariBazaar & 81.3\% & 73.3\% & 73.3\% & 69.3\% & 72.0\% & 73.9\% \\
Haritia Portal & 38.7\% & 68.0\% & 68.0\% & 60.0\% & 56.0\% & 58.1\% \\
RailSetu & 70.7\% & 61.3\% & 65.3\% & 60.0\% & 62.7\% & 64.0\% \\
Selam Airways & 72.0\% & 74.7\% & 74.7\% & 54.7\% & 80.0\% & 71.2\% \\
\bottomrule
\end{tabular}
\label{tab:gpt6luna-popups-site-language}
\end{table*}

% Llama-3.1-8B: all six websites 437/2250 = 19.4%
\begin{table*}[t]
\centering
\small
\setlength{\tabcolsep}{12pt}
\caption{Llama-3.1-8B success rates with interface disruptions for each website and language.}
\begin{tabular}{lccccc c}
\toprule
\textbf{Website} & \textbf{EN} & \textbf{BN} & \textbf{HI} & \textbf{UR} & \textbf{AM} & \textbf{Overall} \\
\midrule
Ministore & 18.7\% & 22.7\% & 22.7\% & 14.7\% & 0.0\% & 15.7\% \\
Redora & 33.3\% & 22.7\% & 24.0\% & 16.0\% & 0.0\% & 19.2\% \\
GariBazaar & 32.0\% & 17.3\% & 21.3\% & 14.7\% & 0.0\% & 17.1\% \\
Haritia Portal & 46.7\% & 25.3\% & 29.3\% & 22.7\% & 1.3\% & 25.1\% \\
RailSetu & 34.7\% & 17.3\% & 25.3\% & 21.3\% & 1.3\% & 20.0\% \\
Selam Airways & 36.0\% & 14.7\% & 28.0\% & 18.7\% & 0.0\% & 19.5\% \\
\bottomrule
\end{tabular}
\label{tab:llama8b-popups-site-language}
\end{table*}

% Gemma-3-4B: all six websites 235/2250 = 10.4%
\begin{table*}[t]
\centering
\small
\setlength{\tabcolsep}{12pt}
\caption{Gemma-3-4B success rates with interface disruptions for each website and language.}
\begin{tabular}{lccccc c}
\toprule
\textbf{Website} & \textbf{EN} & \textbf{BN} & \textbf{HI} & \textbf{UR} & \textbf{AM} & \textbf{Overall} \\
\midrule
Ministore & 13.3\% & 16.0\% & 14.7\% & 13.3\% & 14.7\% & 14.4\% \\
Redora & 2.7\% & 6.7\% & 5.3\% & 6.7\% & 2.7\% & 4.8\% \\
GariBazaar & 12.0\% & 8.0\% & 8.0\% & 12.0\% & 6.7\% & 9.3\% \\
Haritia Portal & 16.0\% & 13.3\% & 17.3\% & 12.0\% & 8.0\% & 13.3\% \\
RailSetu & 12.0\% & 12.0\% & 12.0\% & 12.0\% & 10.7\% & 11.7\% \\
Selam Airways & 10.7\% & 10.7\% & 9.3\% & 6.7\% & 8.0\% & 9.1\% \\
\bottomrule
\end{tabular}
\label{tab:gemma4b-popups-site-language}
\end{table*}

% Gemma-3-12B: all six websites 359/2250 = 16.0%
\begin{table*}[t]
\centering
\small
\setlength{\tabcolsep}{12pt}
\caption{Gemma-3-12B success rates with interface disruptions for each website and language.}
\begin{tabular}{lccccc c}
\toprule
\textbf{Website} & \textbf{EN} & \textbf{BN} & \textbf{HI} & \textbf{UR} & \textbf{AM} & \textbf{Overall} \\
\midrule
Ministore & 29.3\% & 13.3\% & 16.0\% & 6.7\% & 30.7\% & 19.2\% \\
Redora & 5.3\% & 14.7\% & 9.3\% & 9.3\% & 16.0\% & 10.9\% \\
GariBazaar & 12.0\% & 10.7\% & 9.3\% & 5.3\% & 16.0\% & 10.7\% \\
Haritia Portal & 13.3\% & 26.7\% & 25.3\% & 17.3\% & 20.0\% & 20.5\% \\
RailSetu & 14.7\% & 13.3\% & 9.3\% & 12.0\% & 22.7\% & 14.4\% \\
Selam Airways & 14.7\% & 22.7\% & 17.3\% & 14.7\% & 30.7\% & 20.0\% \\
\bottomrule
\end{tabular}
\label{tab:gemma12b-popups-site-language}
\end{table*}

% Gemma-3-27B: all six websites 356/2250 = 15.8%
\begin{table*}[t]
\centering
\small
\setlength{\tabcolsep}{12pt}
\caption{Gemma-3-27B success rates with interface disruptions for each website and language.}
\begin{tabular}{lccccc c}
\toprule
\textbf{Website} & \textbf{EN} & \textbf{BN} & \textbf{HI} & \textbf{UR} & \textbf{AM} & \textbf{Overall} \\
\midrule
Ministore & 14.7\% & 24.0\% & 20.0\% & 20.0\% & 21.3\% & 20.0\% \\
Redora & 12.0\% & 9.3\% & 8.0\% & 9.3\% & 10.7\% & 9.9\% \\
GariBazaar & 12.0\% & 13.3\% & 13.3\% & 17.3\% & 10.7\% & 13.3\% \\
Haritia Portal & 16.0\% & 14.7\% & 14.7\% & 12.0\% & 13.3\% & 14.1\% \\
RailSetu & 13.3\% & 17.3\% & 14.7\% & 14.7\% & 13.3\% & 14.7\% \\
Selam Airways & 17.3\% & 26.7\% & 24.0\% & 20.0\% & 26.7\% & 22.9\% \\
\bottomrule
\end{tabular}
\label{tab:gemma27b-popups-site-language}
\end{table*}

% GPT-OSS-20B: all six websites 550/2250 = 24.4%
\begin{table*}[t]
\centering
\small
\setlength{\tabcolsep}{12pt}
\caption{GPT-OSS-20B success rates with interface disruptions for each website and language.}
\begin{tabular}{lccccc c}
\toprule
\textbf{Website} & \textbf{EN} & \textbf{BN} & \textbf{HI} & \textbf{UR} & \textbf{AM} & \textbf{Overall} \\
\midrule
Ministore & 25.3\% & 21.3\% & 32.0\% & 24.0\% & 14.7\% & 23.5\% \\
Redora & 24.0\% & 16.0\% & 20.0\% & 12.0\% & 9.3\% & 16.3\% \\
GariBazaar & 37.3\% & 28.0\% & 20.0\% & 26.7\% & 8.0\% & 24.0\% \\
Haritia Portal & 28.0\% & 28.0\% & 22.7\% & 29.3\% & 8.0\% & 23.2\% \\
RailSetu & 44.0\% & 36.0\% & 26.7\% & 28.0\% & 12.0\% & 29.3\% \\
Selam Airways & 38.7\% & 38.7\% & 32.0\% & 32.0\% & 10.7\% & 30.4\% \\
\bottomrule
\end{tabular}
\label{tab:gptoss20b-popups-site-language}
\end{table*}

\end{document}